\documentclass[10pt,twocolumn,letterpaper]{article}

\usepackage{cvpr}
\usepackage{times}
\usepackage{epsfig}
\usepackage{graphicx}
\usepackage{amsmath}
\usepackage{amssymb}

\usepackage{enumitem}
\usepackage{dblfloatfix}
\usepackage{multirow}

\usepackage[pagebackref=true,breaklinks=true,letterpaper=true,colorlinks,bookmarks=false]{hyperref}

\cvprfinalcopy % *** Uncomment this line for the final submission

\def\cvprPaperID{2561} % *** Enter the CVPR Paper ID here

\ifcvprfinal\fi
\begin{document}

\definecolor{green}{RGB}{0,100,0}
\definecolor{red}{RGB}{139,0,0}

\def\eg{\emph{e.g. }}
\def\ie{\emph{i.e. }}
\def\Eg{\emph{E.g. }}
\def\etal{\emph{et al. }}

%%%%%%%%% TITLE
\title{Manifestation of Image Contrast in Deep Networks}

\author{
  Arash Akbarinia \quad $\&$ \quad
  Karl R. Gegenfurtner \\
  Abteilung Allgemeine Psychologie \\
  Justus-Liebig-Universit\"at \\
  Giessen, Germany \\
  \texttt{$\lbrace$Arash.Akbarinia, Karl.R.Gegenfurtner$\rbrace$@psychol.uni-giessen.de} \\
}

\maketitle
%\thispagestyle{empty}

%%%%%%%%% ABSTRACT
\begin{abstract}
Contrast is subject to dramatic changes across the visual field, depending on the source of light and scene configurations. Hence, the human visual system has evolved to be more sensitive to contrast than absolute luminance. 
% Luminance changes much more than contrast, that's why we are sensitive to contrast rather than luminance. Shouldn't the first word be "Luminance" then?
This feature is equally desired for machine vision: the ability to recognise patterns even when aspects of them are transformed due to variation in local and global contrast. In this work, we thoroughly investigate the impact of image contrast on prominent deep convolutional networks, both during the training and testing phase. The results of conducted experiments testify to an evident deterioration in the accuracy of all state-of-the-art networks at low-contrast images. We demonstrate that ``contrast-augmentation" is a sufficient condition to endow a network with invariance to contrast. This practice shows no negative side effects, quite the contrary, it might allow a model to refrain from other illuminance related over-fittings. This ability can also be achieved by a short fine-tuning procedure, which opens new lines of investigation on mechanisms involved in two networks whose weights are over 99.9\% correlated, yet astonishingly produce utterly different outcomes. Our further analysis suggest that the optimisation algorithm is an influential factor, however with a significantly lower effect; and while the choice of an architecture manifests a negligible impact on this phenomenon, the first layers appear to be more critical.
\end{abstract}

%%%%%%%%% BODY TEXT
\section{Introduction}

In visual perception, contrast is defined as the difference in brightness of an object with its surroundings. 
% It is not really defined as the difference. It is the difference divided by the average brightness, and this makes it fairly invariant to lighting changes. A pure contrast reduction would occur in fog, for example.
This is primarily determined by the source of light. The illumining lux at bright sunlight is larger than starlight by a factor of one billion \cite{schlyter2009radiometry}. Subsequently, contrast is higher to a greater extent in a well illuminated scene in comparison to a dim environment. In addition to this global factor, local configurations of a scene's constituents cause considerable variation in contrast of present objects by creating shadows and reflections \cite{frazor2006local}.

\begin{figure}[ht]
    \centering
    \setlength{\tabcolsep}{2pt}
    \begin{tabular}{cc}
        \includegraphics[width=0.41\columnwidth]{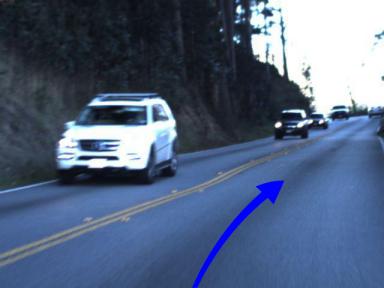} & \includegraphics[width=0.41\columnwidth]{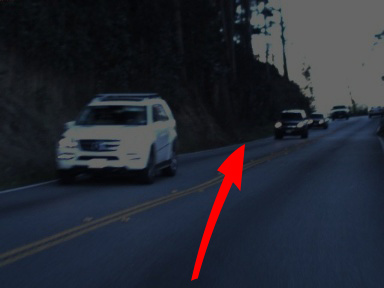} \\
        100\% & 40\%  \\
    \end{tabular}
    \caption{A sample synthetic image used to test autonomous driving. The system exhibits erroneous behaviour at 40\% image contrast and depicted by the red arrow. Image credit \url{https://deeplearningtest.github.io/deepTest/}.}
    \label{fig:adex}
\end{figure}

Although we all prefer a colourful day, our visual perception is not impaired in faint contrast. 
% Under dim light (in the dark) vision is impaired. Acuity is poor in rod vision. And in your experimenst below you do not change light level. Mean luminance is constant.
This is expected from an artificially intelligent machine as well. Autonomous driving cars have been reported to cause fatal accidents in low-contrast visibility conditions \cite{tian2018deeptest} (see Figure \ref{fig:adex}). This emphasises the importance of designing a \textit{contrast-invariant} machine vision. Previous works in the literature have reported that the accuracy of most prominent deep networks at object classification falls noticeably at low-contrast images \cite{geirhos2017comparing, dodge2016understanding, akbarinia2018contrast}. Therefore, further investigation is required to address, at least, two fundamental research questions around this topic:
\begin{enumerate}[label=\roman*]%[noitemsep,nolistsep]
    \item \textbf{How} a contrast-invariant network can be achieved. What is the most influential factor, \ie the architecture of a model, the training procedure, or a combination of both?
    \item \textbf{What} sorts of mechanism a network learns in order to prevent variation of contrast in input image propagating to its output. Is it a specific kernel, a layer, or an interaction among them?
\end{enumerate}
Although both questions are equally important, it is worth acknowledging that a resolution to the former does not necessarily reveal an explanation to the latter. For instance, perhaps networks undergone a specific training procedure produce a same set of outputs across different levels of contrast, nevertheless this leaves us with little insight about the exact mechanism in terms of neuronal operations that leads to this feature.

Preceding related investigations have been mainly focused on evaluating the performance of the deep neural networks (DNN) exposed to poor visual information. One study measured classification accuracy of four networks on \textit{ImageNet} data set under five image distortions, including contrast reduction \cite{dodge2016understanding}. Similarly, \cite{geirhos2017comparing} compared the ability of three DNNs to human observers in a psychophysical experiment, in which quality of images was degraded. Results of both studies indicate that while some networks perform substantially better than others, all decay at low-contrast images (\ie about 20\% level of contrast).

Hitherto, the second question has not been inquired. This was addressed in a recent study \cite{akbarinia2018contrast} that analysed activation map of kernels in eight DNNs. Their findings suggest that relative position of the first max-pooling is an influential factor for a network to accomplish invariance to contrast. Their methodology is in harmony with the rationale to dissect a network into human-interpretable concepts \cite{bau2017network}. However, networks examined by~\cite{akbarinia2018contrast} were state-of-the-art models that were trained with distinct procedures. Therefore, their findings might be inconclusive in disentanglement of contrast representation in a neural networks.

In this article we attempt to contribute to both questions. The principal capacity that deep learning has provided for computational models is to learn intricate patterns from large data sets \cite{lecun2015deep}, therefore, data augmentation has been actively used as a tool to increase robustness to common image transformations and avoid over-fitting \cite{krizhevsky2012imagenet,simonyan2014very}. Along this line, we investigated the impact of \textit{contrast-augmentation} on training and fine-tuning DNNs. The results of our experiments show that this is an effective approach to make networks invariant to image contrast even at extreme low levels, with no side effects. 

Comprehending feature representation and its disentanglement is of great interest to the research community \cite{bengio2013representation}. Consequently, in order to gain insights about the features learnt by DNNs, a large body of literature has developed techniques to visualise its internal units (\eg \cite{zeiler2014visualizing,mahendran2015understanding}). Others have proposed a binary segmentation task to study every neuron of a network~\cite{bau2017network}. These methods are primarily applicable to high level perceptual concepts (\eg objects) or tangible low level features (\eg colours). In contrary, contrast is a fundamental visual feature in natural scenes, independent of luminance \cite{mante2005independence}, thus, it is cumbersome to decipher its role through above-stated approaches. Instead, we tackled the second question from two other angles:
\begin{itemize}
    \item We trained various networks under a completely controlled procedure: forcing all variables to be identical except one to be singled out at each experiment. This allows us to identify the influence of each factor. Additionally, we compared the learnt weights in a pair of twin networks to localise notion of contrast in an architecture. Previous studies suggest a concept is often encoded with a combination of several neurons \cite{agrawal2014analyzing, gonzalez2018semantic}.
    
    \item Analysing all neurons of the same network under multiple levels of contrast. We compared activation of kernels between inputs correctly classified at lower image contrast to the misclassified instances. This is a common practice in neuroscience due to limit of access to a large set of brains \cite{kandel2000principles}. The hypothesis is if a specific kernel or layer represents the concept of contrast, this would be revealed by comparing the activity of the same network under successful and failure trials.
\end{itemize}

Our analysis suggest that the first few layers of a network are protagonists of a mechanism to achieve invariance to image contrast. However, we did not discover any individual kernel, layer, or a combination of both that is capable of explaining how two networks whose weights are 99.9\% correlated produce utterly different results. % Yup, that's the big question we should work on!

\section{Methodology}

\subsection{Setup}

\subsubsection{Data set}

We conducted our experiments on \textit{ImageNet} data set \cite{krizhevsky2012imagenet} which is collection of one thousand object categories. The training-set contains 1.3 million images (\ie 1300 per category) and the validation-set consists of 50 thousands images (\ie 50 per category).

\subsubsection{Networks}

We studied thirteen distinguished networks in state-of-the-art: \textit{VGG16} \cite{simonyan2014very}, \textit{VGG19} \cite{simonyan2014very}, \textit{ResNet50} \cite{he2016deep}, \textit{InceptionV3}~\cite{szegedy2016rethinking}, \textit{Xception} \cite{chollet2017xception}, \textit{InceptionResNetV2} \cite{szegedy2017inception}, \textit{MobileNet} \cite{howard2017mobilenets}, \textit{MobileNetV2} \cite{sandler2018mobilenetv2}, \textit{DenseNet121} \cite{huang2017densely}, \textit{DenseNet169} \cite{huang2017densely}, \textit{DenseNet201} \cite{huang2017densely}, \textit{NASNetMobile} \cite{zoph2017learning}, and \textit{NASNetLarge}~\cite{zoph2017learning}.

The weights of all these models were obtained from Keras platform\footnote{\url{https://keras.io/applications/}}. It is worth emphasising that each of the aforementioned networks have been trained on \textit{ImageNet} data set, however each pretrained network has experienced a different procedure: such as in the choice of optimiser, number of epochs, or types of image augmentation.

\subsubsection{Contrast manipulation}

There are many possible formulations to change contrast of an image. One approach would be to blend the input image with a grey image \cite{dodge2016understanding}. Another is to modulate Michelson contrast \cite{michelson1995studies}. We opted for the latter through this equation:
\begin{align}
I_c(x,y) = \frac{c}{100} \times I(x,y) + \frac{1 - \frac{c}{100}}{2},
\label{eq:contrast}
\end{align}
where $I$ is the input image, $\left\{x,y\right\}$ are pixel coordinates and $c$ is the contrast level. In this way, our findings are comparable to previous studies on this topic \cite{geirhos2017comparing,akbarinia2018contrast}.

\subsection{Experiments}

\subsubsection{Evaluating pretrained networks}
\begin{figure}[ht]
    \centering
    \setlength{\tabcolsep}{2pt}
    \begin{tabular}{cccc}
        \includegraphics[width=0.235\columnwidth]{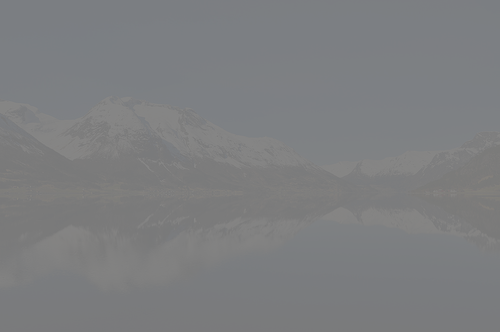} & \includegraphics[width=0.235\columnwidth]{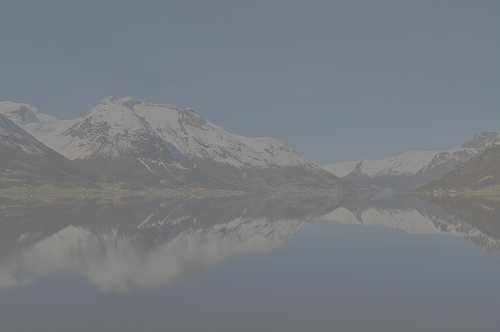} & \includegraphics[width=0.235\columnwidth]{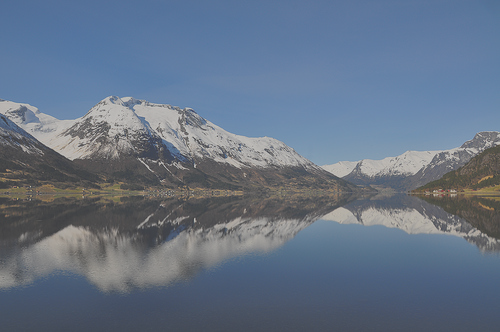} & \includegraphics[width=0.235\columnwidth]{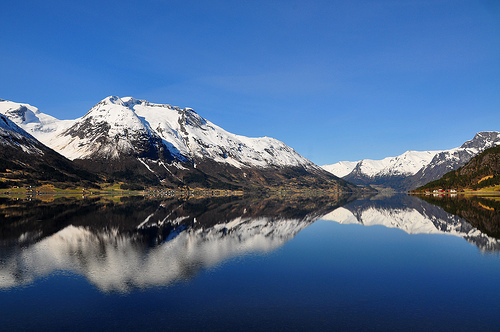} \\
        5\% & 15\% & 50\% & 100\% \\
    \end{tabular}
    \caption{An exemplary image under different levels of contrast.}
    \label{fig:imageexample}
\end{figure}

We evaluated the classification accuracy of each of those thirteen prominent pretrained networks on the validation-set of \textit{ImageNet} data set under seven levels of contrast: namely $c \in \left\{ 1, 5, 15, 30, 50, 75, 100 \right\} \%$ in Eq.~\ref{eq:contrast}. See Figure~\ref{fig:imageexample} for an exemplary input image.

\noindent
Prior to feeding a network with an image:
\begin{itemize}%[noitemsep,nolistsep]
    \item First, it was resized to its smaller edge and the central square was cropped according to the input size of each network (\ie \textit{NASNetLarge} receives images of size $331 \times 331$, \textit{Xception}, \textit{InceptionV3}, and \textit{InceptionResNetV2} $299 \times 299$, and all the rest $224 \times 224$).
    \item Second, it was preprocessed with the same function utilised during the original training procedure of that network.
\end{itemize}

\subsubsection{Training networks from scratch}

The experiment described above sheds light on state-of-the-art in image classification across different levels of contrast. However, each model is of a completely distinct nature (\eg different architectures, number of parameters, type of optimisation, the objective function, training procedure, preprocessing function \etc). Therefore, studying pretrained DNNs alone does not allow us to disentangle the impact of each factor to the role of image contrast, and subsequently to discover the mechanism a network has learnt to become invariant to this feature. In other words to generalise along its dimension.

Accordingly, we studied three of those networks (\textit{InceptionV3}, \textit{ResNet50} and \textit{DenseNet201}) in a greater details by means of training various instances of them from scratch under identical conditions. We chose these models because: (i) their architectures consist of a similar number of parameters, (ii) their classification accuracy at 100\% image contrast is comparable, (iii) their accuracy at lower levels of contrast is very different.

We investigated four factors: (i) whether a specific architecture results in invariance to image contrast, (ii) effect of relative position of the first max-pooling \cite{akbarinia2018contrast}, (iii) choice of the optimisation algorithm, and (iv) exposure to multiple levels of contrast during the training phase.

We trained each instance of these networks in ten epochs with a batch size of 32 on a single GPU. In this experiment, we excluded all standard ``augmentation" (\eg flipping, random cropping, scaling, \etc) in our training phase, due to the random nature of these procedures that makes the comparison more complicated and less accurate.

\subsubsection{Fine-tuning pretrained networks}

We selected five of those pretrained networks with the criteria to cover various performances proportional to their number of parameters (\ie \textit{MobileNetV2}, \textit{ResNet50}, \textit{VGG16}, \textit{NASNetMobile}, and \textit{InceptionV3}) and fine-tuned their weights with ``\textit{contrast-augmentation}". This procedure is not augmentation in the sense of increasing the number of training images. It merely refers to the manipulation of the image contrast during the training with a random contrast level in the range of 1 to 100\% according to Eq.~\ref{eq:contrast}. Therefore, each network is essentially exposed to the same number of training images at each epoch.

Irrespective of the optimisation configurations of a network at its original training procedure, we fine-tuned all instances with an Adam optimiser \cite{kingma2014adam} and a categorical cross-entropy objective function. We set the learning rate and decay parameters both equal to $10^{-6}$ for all our experiments. This fine-tuning consists of five epochs of retraining with ``\textit{contrast-augmentation}" with a batch size of 32 on a single GPU. During the fine-tuning, we included the following standard ``augmentation" procedures: \ie random horizontal flipping, zooming (within a 20\% scale), and shifting (within a 20\% range).

\section{Results\protect\footnote{Source code and experimental materials are available at \url{https://goo.gl/GkdZQt}.}}

\subsection{Pretrained networks}

The top-1 classification accuracy of all examined networks on the validation-set of \textit{ImageNet} at different levels of contrast are illustrated in Figure \ref{fig:prominents}. It is important to emphasise that each curve is divided by its value at 100\% image contrast to facilitate the comparison. The objective of this experiment is not to evaluate the absolute accuracy of each network, but rather to investigate which network can retain its performance at lower contrast levels. The absolute accuracy (top-1 and top-5) of fine-tuned networks are reported in Table \ref{tab:all}.

\begin{figure*}[ht]
    \centering
    \begin{tabular}{cc}
         \includegraphics[width=\columnwidth]{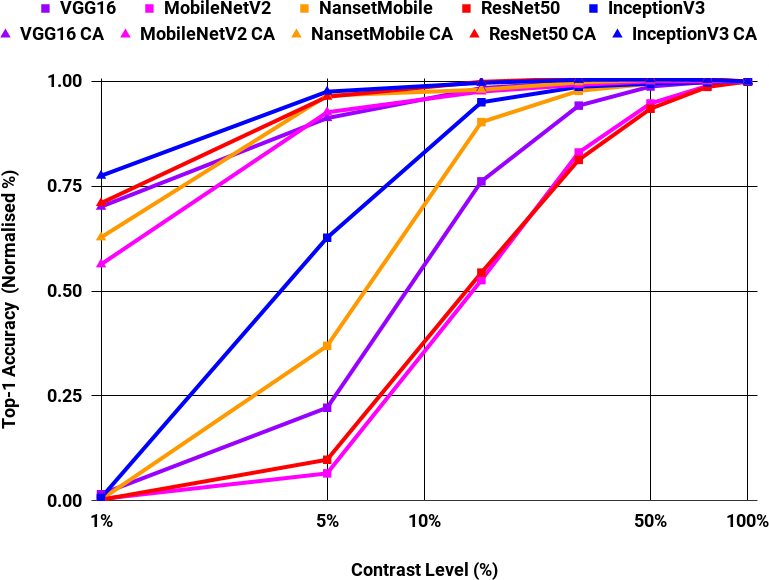} & \includegraphics[width=\columnwidth]{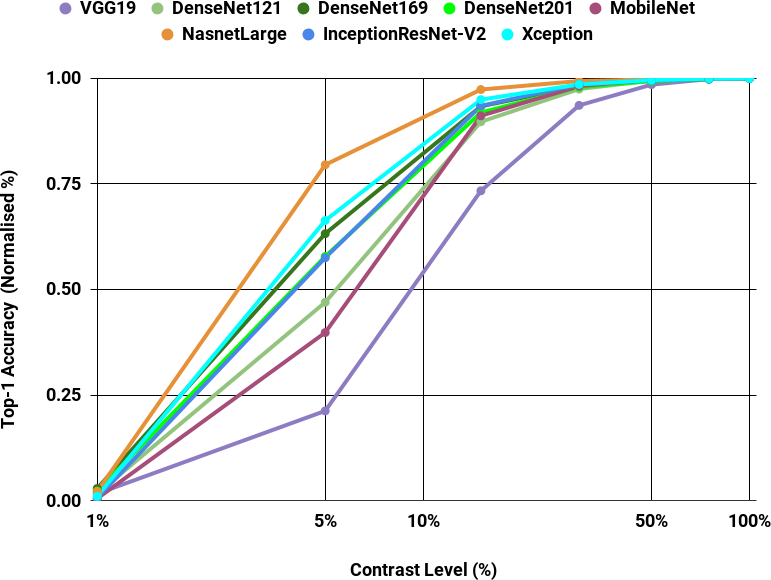} 
    \end{tabular}
    \caption{The top-1 classification accuracy of various networks on the validation-set of \textit{ImageNet} data set. The curves are \textbf{normalised} to perfect accuracy on 100\% level of contrast. On the left panel: those curves with a \textbf{triangle} shape have gone through \textit{contrast-augmented} fine-tuning, initialised with the weights of the \textbf{square} shaped curves. On the right panel: all the rest of the pretrained models obtained from Keras platform. Interested readers are encouraged to refer to supplementary materials for a similar figure of top-5 classification accuracy.}
    \label{fig:prominents}
\end{figure*}

It is evident from Figure \ref{fig:prominents} that all networks retain their peak performance at 75\% level of contrast. After that, the accuracy of \textit{ResNet50} and \textit{MobileNetV2} deteriorates at a sharper rate (note the red and magenta curves at the bottom of the left panel) followed by \textit{VGG16} and \textit{VGG19} (note the purple curves in the middle of both panels). Others perform very well down till 30\% level of contrast. However, they experience a large drop by the time image contrast reaches 15 and 5\%. And finally, at 1\% level of contrast all pretrained networks are at chance level.

Contrary to this, all the five \textit{contrast-augmented} networks retain almost perfectly their peak accuracy down till 5\% image contrast, and at 1\% level of contrast, on average, they score about 70\% of their original performance (note the triangle shaped curves at the top of the left panel in Figure~\ref{fig:prominents}). The difference among fine-tuned networks at 1\% image contrast could be due to the number of parameters a model is consist of (\ie \textit{MobileNetV2} and \textit{NASNetMobile} are in the order of five millions parameters, substantially less than the others).

\begin{table*}[hb]
    \centering
    \begin{tabular}{|l|l|ccccc|ccccc|}
        \cline{3-12}
        \multicolumn{2}{l|}{} & \multicolumn{5}{c|}{\textbf{Top-1}} & \multicolumn{5}{c|}{\textbf{Top-5}}\\ \cline{2-12}
        \multicolumn{1}{l|}{} & \textbf{Contrast level} & \textbf{1\%} & \textbf{5\%} & \textbf{15\%} & \textbf{50\%} & \textbf{100\%} & \textbf{1\%} & \textbf{5\%} & \textbf{15\%} & \textbf{50\%} & \textbf{100\%} \\ \hline
        \multirow{2}{*}{\textbf{\textit{InceptionV3}}} & Original & \textcolor{red}{0.00} & \textcolor{red}{0.49} & \textcolor{red}{0.74} & 0.77 & 0.77 & \textcolor{red}{0.17} & \textcolor{red}{0.72} & \textcolor{red}{0.91} & 0.93 & 0.94 \\ \cline{2-12}
         & Fine-tuned & \textcolor{green}{0.60} & \textcolor{green}{0.75} & \textcolor{green}{0.77} & 0.77 & 0.77 & \textcolor{green}{0.82} & \textcolor{green}{0.93} & \textcolor{green}{0.94} & 0.94 & 0.94 \\ \hline \hline
        \multirow{2}{*}{\textbf{\textit{ResNet50}}} & Original & \textcolor{red}{0.00} & \textcolor{red}{0.07} & \textcolor{red}{0.40} & \textcolor{red}{0.69} & 0.74 & \textcolor{red}{0.01} & \textcolor{red}{0.17} & \textcolor{red}{0.65} & \textcolor{red}{0.89} & 0.92 \\ \cline{2-12}
         & Fine-tuned & \textcolor{green}{0.52} & \textcolor{green}{0.71} & \textcolor{green}{0.74} & \textcolor{green}{0.74} & 0.74 & \textcolor{green}{0.76} & \textcolor{green}{0.90} & \textcolor{green}{0.92} & \textcolor{green}{0.93} & 0.92 \\ \hline \hline
        \multirow{2}{*}{\textbf{\textit{VGG16}}} & Original & \textcolor{red}{0.01} & \textcolor{red}{0.15} & \textcolor{red}{0.53} & 0.69 & 0.70 & \textcolor{red}{0.03} & \textcolor{red}{0.36} & \textcolor{red}{0.78} & 0.89 & 0.89 \\ \cline{2-12}
         & Fine-tuned & \textcolor{green}{0.50} & \textcolor{green}{0.65} & \textcolor{green}{0.70} & 0.71 & 0.71 & \textcolor{green}{0.70} & \textcolor{green}{0.87} & \textcolor{green}{0.90} & 0.90 & 0.90 \\ \hline \hline
        \multirow{2}{*}{\textbf{\textit{NASNetMobile}}} & Original & \textcolor{red}{0.00} & \textcolor{red}{0.27} & \textcolor{red}{0.66} & 0.73 & 0.73 & \textcolor{red}{0.08} & \textcolor{red}{0.48} & \textcolor{red}{0.86} & 0.91 & 0.91 \\ \cline{2-12}
         & Fine-tuned & \textcolor{green}{0.45} & \textcolor{green}{0.70} & \textcolor{green}{0.71} & 0.73 & 0.73 & \textcolor{green}{0.59} & \textcolor{green}{0.88} & \textcolor{green}{0.90} & 0.91 & 0.91 \\ \hline \hline
        \multirow{2}{*}{\textbf{\textit{MobileNetV2}}} & Original & \textcolor{red}{0.00} & \textcolor{red}{0.05} & \textcolor{red}{0.38} & \textcolor{red}{0.68} & 0.71 & \textcolor{red}{0.08} & \textcolor{red}{0.12} & \textcolor{red}{0.64} & \textcolor{red}{0.88} & 0.90 \\ \cline{2-12}
         & Fine-tuned & \textcolor{green}{0.40} & \textcolor{green}{0.66} & \textcolor{green}{0.69} & \textcolor{green}{0.71} & 0.71 & \textcolor{green}{0.55} & \textcolor{green}{0.88} & \textcolor{green}{0.90} & \textcolor{green}{0.90} & 0.90 \\ \hline
    \end{tabular}
    \caption{The classification accuracy comparison of various networks with their fine-tuned \textit{contrast-augmented} offspring.}
\label{tab:all}
\end{table*}

\subsection{Networks trained from scratch}

The top-1 classification accuracy of networks trained from scratch on validation-set of \textit{ImageNet} at different levels of contrast are illustrated in Figure \ref{fig:trained}. It is important to emphasise that these curves, similar to those in Figure \ref{fig:prominents}, are normalised to obtain perfect accuracy at full image contrast. Furthermore, let us remind ourselves that these networks have been trained only for ten epochs with no image augmentation. This explains why their overall performance is lower in comparison to the pretrained networks reported in Figure~\ref{fig:prominents}. 

In order to accurately identify influential factors on image contrast, we controlled every aspect of the training procedures to make them as identical as possible for all networks: \eg preprocessing functions, weights initialisation, no shuffling in the order of images, batch size, and input size. No image augmentation was used except for those labelled as ``contrast-augmented" in which the contrast of training images were randomly adjusted within the range of 1 to 100\%.

We trained an instance of each architecture with two types of optimisation algorithm: Adam (learning rate set to $10^{-3}$ and decay is equal to $10^{-6}$) and SGD (learning rate set to $10^{-1}$ and decay is equal to $10^{-4}$). We used categorical cross-entropy as the objective function for all our experiments.

Within each class of architecture, all instances have approximately the same number of parameters. The ``Area1'' labels refers to the distribution of convolutional kernels prior to the first max-pooling layer. ``Area1\_1'' means there is one convolutional layer before the first max-pooling. ``Area1\_2'' and ``Area1\_3'' refer to two and three convolutional layers, correspondingly.

There are three phenomena emerging from Figure~\ref{fig:trained} that are worth to be highlighted:
\begin{enumerate}[label=\roman*]
    \item All contrast-augmented networks retain their peak performance perfectly down till 5\% image contrast; and at 1\% level of contrast they obtain, on average, half of their accuracy at full contrast (see the triangle shaped curves at the top of the figure lying on the line $y=1$).
    \item Among the rest, those trained with Adam optimiser perform better at lower levels of contrast in comparison to their identical twins trained with SGD. This is true irrespective of the network architecture (compare the square shaped curves to the circle ones).
    \item Overall, the architecture of a network appears to be of minor importance, since \textit{InceptionV3}, \textit{ResNet50} and \textit{DenseNet201} produce comparable results under identical conditions. However, a tiny anecdote appears within each family of the networks: those with multiple convolutional layers prior to the first max-pooling perform slightly better in comparison to their counterparts with a single convolutional layer \cite{akbarinia2018contrast} (\eg compare the dark and light squares or circles).
\end{enumerate}

\begin{figure}[ht]
    \centering
    \includegraphics[width=\columnwidth]{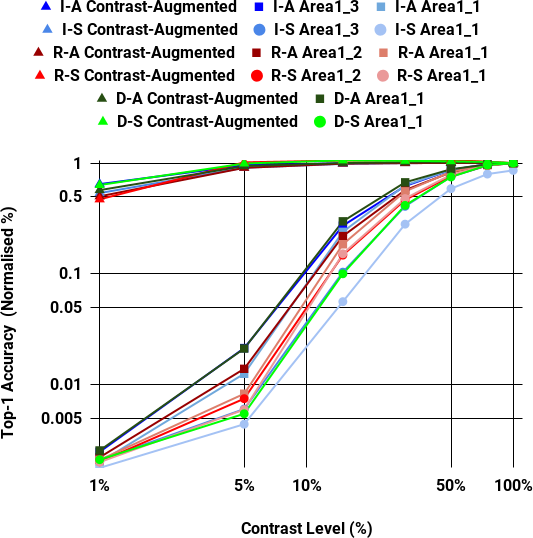}
    \caption{The top-1 classification accuracy of various networks on the validation-set of \textit{ImageNet}. The curves are \textbf{normalised} to perfect accuracy on 100\% level of contrast. Each legend starts with an \textbf{abbreviation}, the first corresponds to network (\textit{I: InceptionV3} -- \textbf{Blue}, \textit{R: ResNet50} -- \textbf{Red}, \textit{D: DenseNet201} -- \textbf{Green}) and the second to optimiser (\textit{A: Adam} -- \textbf{Square}, \textit{S: SGD} -- \textbf{Circle}). Those with a \textbf{triangle} shape have gone through a \textit{contrast-augmented} training procedure. Interested readers are encouraged to refer to supplementary materials for top-5 classification accuracy.}
    \label{fig:trained}
\end{figure}

\section{Discussion}

\subsection{The how question}
The results of the fine-tuning experiment reported in Figure \ref{fig:prominents} and Table \ref{tab:all} indicate three significant phenomena:
\begin{enumerate}[label=\roman*]
    \item Pretrained networks in state-of-the-art, irrespective of their architecture or original training procedure, can become almost perfectly invariant to image contrast with a few epochs of contrast-augmented fine-tuning (compare the triangle shaped curves in the left panel of Figure \ref{fig:prominents} to the square ones).
    
    Most networks experience a dramatic change in their performance across multiple levels of contrast. For instance, the original \textit{ResNet50} barely retains 10\% of its peak performance at 5\% image contrast. However, the fine-tuned offspring maintains 96\% of its accuracy under the same condition (see Figure \ref{fig:resexample} for an example).
      
    \item An obvious objection could be raised that the contrast-augmented fine-tuning would harm the absolute accuracy of a network at 100\% level of contrast. The verdict is no, as it is evident in Table \ref{tab:all}. All fine-tuned networks match or exceed their original accuracy at full contrast image, till the second decimal point (for both measures of top-1 and top5). 
    
    At the same time, the gain is striking for all networks at many lower image contrasts. Compare red and green figures in Table \ref{tab:all}. To name one as an example, the original \textit{ResNet50} obtains 0 and 7\% classification accuracy at contrast levels 1 and 5\%, while the fine-tuned version scores 52 and 71\%, respectively.

\begin{figure}[ht]
    \centering
    \setlength{\tabcolsep}{2pt}
    \begin{tabular}{cccc}
        \includegraphics[width=0.235\columnwidth]{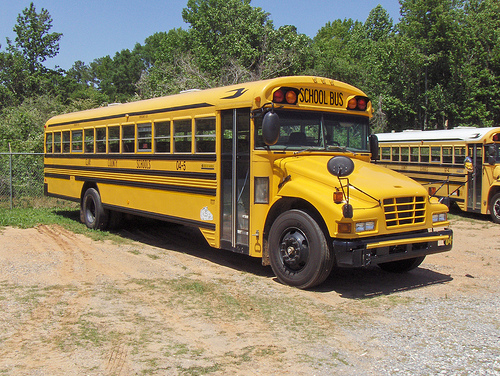} & \includegraphics[width=0.235\columnwidth]{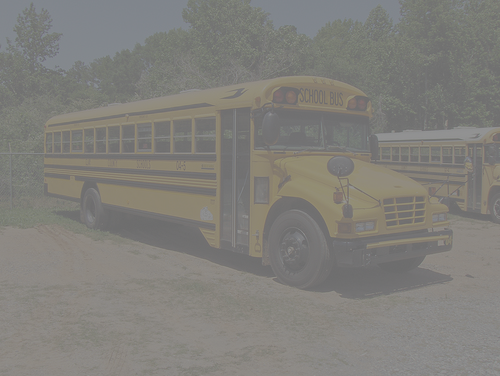} & \includegraphics[width=0.235\columnwidth]{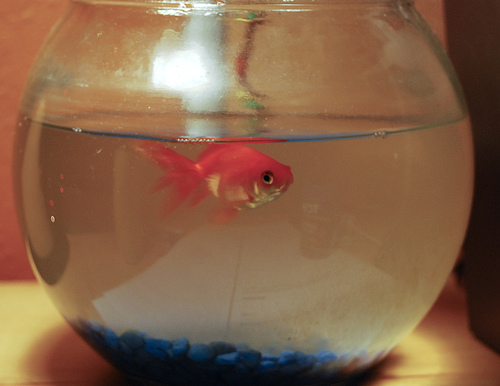} & \includegraphics[width=0.235\columnwidth]{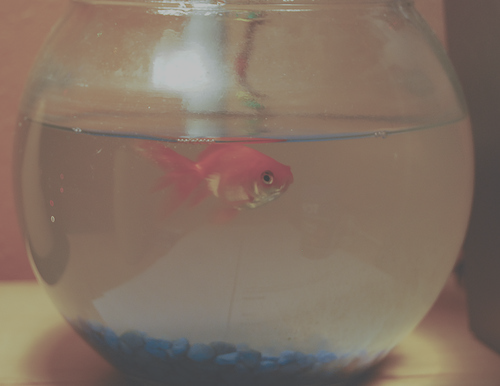} \\
        100\% & 15\% & 100\% & 50\% \\
    \end{tabular}
    \caption{Two examples in which at 100\% image contrast both the original \textit{ResNet50} and its fine-tuned offspring correctly classify the present objects. At a lower level of contrast the original model fails while the fine-tuned version succeeds.}
    \label{fig:resexample}
\end{figure}

    \item This boost in performance of object classification for images of reduced contrast, appears to be little influenced by the original accuracy of a network at those conditions. For instance, the original \textit{ResNet50} performs extremely poor at low levels of contrast, while \textit{InceptionV3} performs relatively well in those conditions. Nevertheless, the corresponding contrast-augmented versions of both obtain close to perfect accuracy down till 5\% level of contrast.
\end{enumerate}

The results of the training networks in identical conditions (Figure \ref{fig:trained}) support these findings:
\begin{enumerate}[label=\roman*]
    \item Exposure to multiple image contrasts during the training phase is the crucial element.
    \item The choice of optimisation is a secondary factor, which also technically refers to the training procedure.
    \item The overall impact of a network architecture is diminutive, although lower layers appear somehow pivotal.
\end{enumerate}

These findings suggest that exposure to multiple levels of contrast allows a network to adequately learn a set of parameters to essentially accomplish object classification invariance to image contrast, which can notoriously change at all time \cite{frazor2006local}. This could present potential implications for distinct vision related disciplines:
\begin{itemize}
    \item From a machine vision perspective, one interpretation could be that a network consists of millions of parameters is probably capable of encoding many more features within the same architecture \cite{li2017learning,kokkinos2017ubernet}.
    
    Alternatively, looking at this inversely suggests that it should be possible to achieve the same level of accuracy at a single level of image contrast with smaller networks \cite{iandola2016squeezenet,rastegari2016xnor}.
    
    \item From a visual perception perspective, this empowers the empirical theory of vision, inference through successful behaviour \cite{purves2011we}, by providing more evidence that past experience indeed is an essential part of a visually intelligent system.
\end{itemize}

\subsubsection{The amount of training required}

Analysing the evolution of the fine-tuning procedure in a greater details (see Figure \ref{fig:evolution} for \textit{RestNet50}) suggests that essentially the first epoch is the most influential one, while epochs two to five allow the network to adjust its parameters more adequately to boost the performance at very low levels of image contrast (\ie 1 and 5\%).

\begin{figure}[ht]
    \centering
    \includegraphics[width=\columnwidth]{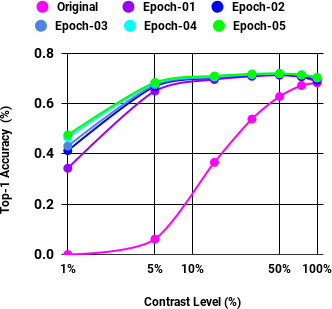}
    \caption{The top-1 classification accuracy of fine-tuned \textit{ResNet50} in various epochs on the validation-set of \textit{ImageNet} data set.}
    \label{fig:evolution}
\end{figure}

\subsubsection{Degradation in other visual information}

It can be reasonably objected that contrast-augmentation could cause undesired side effects on robustness of the network to degradation of other visual features. A contrary school of thought could argue that exposure to random image contrast compels the network to rely on more abstract representation, and consequently less impaired by other image transformations. We scrutinised this opposite views by comparing the performance of a network to its fine-tuned version over validation-set of \textit{ImageNet} under a diverse set of image manipulations (refer to supplementary materials):
\begin{itemize}
    \item Gaussian blurring with square windows of side 3, 9 and 27 pixels. We did not observe any systematical evidence for either side of the argument and essentially the classification accuracy of original networks and their fine-tuned versions were on-par.
    \item Salt \& Pepper noise of 1, 5 and 10\%, or uniformly distributed noise. Similar verdict as Gaussian blurring.
    \item Gamma correction with $\gamma \in \left\{ 0.3, 0.8, 1.2, 3.0 \right\}$. Fine-tuned networks performed substantially better (maximally in the order of 20\%) in case of gamma compression while performing slightly worse (maximally in the order of 2\%) in case of gamma expansion. This suggests that the linear contrast-augmentation has a positive effect on nonlinear contrast manipulation as well.
    \item Varying illumination conditions by multiplying two colour channels at a time with a constant of values 0.25, 0.50, and 0.75 (\ie making the image reddish, greenish, or bluish respectively). We observed that the fine-tuned network perform better at this experiment. This is in line with previously reported importance of contrast in computational colour constancy~\cite{akbarinia2018colour}.
\end{itemize}

The results of these experiments suggest that contrast-augmentation could potentially make DNNs more robust towards other changes that occur in illumination of a scene. This should cause no surprise as contrast aware non-learning algorithms have been incorporated into a wide range of computer vision applications with encouraging results (\eg \cite{akbarinia2017feedback,itti2001computational,perazzi2012saliency}).

%For instance, accuracy of the original \textit{ResNet50} and \textit{VGG16} by 5-6\% if validation-set of ImageNet data set is resized to an image of 224x224 pixels instead of cropping the central part of that size. Contrary to this, the contrast-augmented versions do not suffer from this and perform equally good in both conditions (pruning over-fitting).

%We further examine how interpretability is affected by training data sets, training techniques like dropout [28] and batch normalization [13], and supervision by different primary tasks.

\subsection{The what question}

\subsubsection{Twin networks comparison}
In order to better grasp what is internally altered that allows a model to become invariant to image contrast, first, we compared raw weights of the original networks to their fine-tuned offspring (see Table \ref{tab:metrics}): 
\begin{itemize}
    \item The raw weights are more than 99.9\% linearly correlated with a negligible variation across layers.
    \item The absolute difference between raw weights is in the order of $10^{-4}$. Given that the average range of weights in these networks are typically in the the order of $10^{-2}$, this implies that the absolute difference is merely 1\%.
    \item There is more than 99.9\% mutual dependence between raw weights of the original and fine-tuned networks. 
\end{itemize}

\begin{table}[ht]
    \centering
    \begin{tabular}{|l|ccc|}
        \cline{2-4}
        \multicolumn{1}{l|}{} & \textbf{PCC} & \textbf{Difference} & \textbf{NMI} \\ \hline 
        \textbf{\textit{InceptionV3}} & $0.99$ & $8 (\pm 3) \times10^{-4}$ & $0.99$ \\ \hline \hline
        \textbf{\textit{ResNet50}} & $0.99$ & $10 (\pm 4) \times10^{-4}$ & $0.99$\\ \hline \hline 
        \textbf{\textit{VGG16}} & $0.99$ & $4 (\pm 2) \times10^{-4}$ & $0.99$\\ \hline \hline
        \textbf{\textit{NASNetMobile}} & $0.99$ & $18 (\pm 6) \times10^{-4}$ & $0.99$\\ \hline \hline
        \textbf{\textit{MobileNetV2}} & $0.99$ & $25 (\pm 9) \times10^{-4}$ & $0.99$\\ \hline
    \end{tabular}
    \caption{The comparison of raw weights of an original network to its fine-tuned offspring. Metrics from left to right: Pearson correlation coefficient (\textbf{PCC}), the absolute \textbf{Difference}, and Normalised Mutual Information (\textbf{NMI}). Values are averaged out over all layers of a network. Standard deviations are not shown for PCC and NMI since they were 0 till the third decimal.}
    \label{tab:metrics}
\end{table}

\begin{figure}[ht]
    \centering
    \includegraphics[width=\columnwidth]{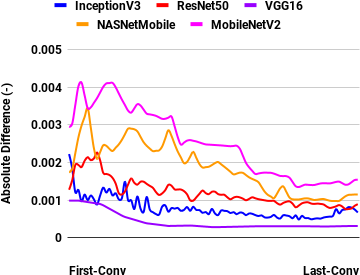}
    \caption{The absolute difference of kernel weights between an original network and its fine-tuned offspring.}
    \label{fig:weights}
\end{figure}

Inspecting the absolute difference and mutual information at each layer separately, exhibits a tendency among all examined networks: first convolutional layers have a smaller mutual information and a larger absolute difference. This relation inverts itself as we progress towards the last convolutional layer. See Figure \ref{fig:weights} for absolute difference and refer to supplementary materials for mutual information. This suggests that the biggest difference between a network and its fine-tuned offspring is at the first few layers. This phenomenon sounds logical, given contrast is a low level visual feature, subsequently, no account of its variation at the start of a feed-forward model would cause a propagation of larger impact to higher layers \cite{lecun2015deep}. This is in accordance with the reported contrast-invariant cells of lower cortical areas in biological vision~\cite{hubel1962receptive}.

In the analysis discussed above, two networks are compared at corresponding layers represented through the average of all their constituent kernels. This assumes a particular layer is entirely capturing contrast of an image. A more realistic alternative could hypothesise that the difference occur at a finer level among individual kernels, whose impact is vanished at the analysis above due to the limitations of an average representation.

One approach to highlight true differences between kernels of both networks is nominal thresholds, similar to the idea of binary classification~\cite{bau2017network}. We examined this by counting the number of kernels pairs whose correlation is outside of  standard variation. No systematic pattern emerged generic to all networks, however we discovered discernible differences at each network. For instance, all ``\textit{branch\_2c}" convolutional layers of \textit{ResNet50} stand out with many constituent kernels having a correlation smaller than $0.75$ between the original network and its fine-tuned version (see blue bars in Figure \ref{fig:r_w_a}). No other layer holds a single kernel of this criteria. It is worth reminding ourselves that the average correlation among all kernels is greater than $0.999$.

\begin{figure}[ht]
    \centering
    \includegraphics[width=\columnwidth]{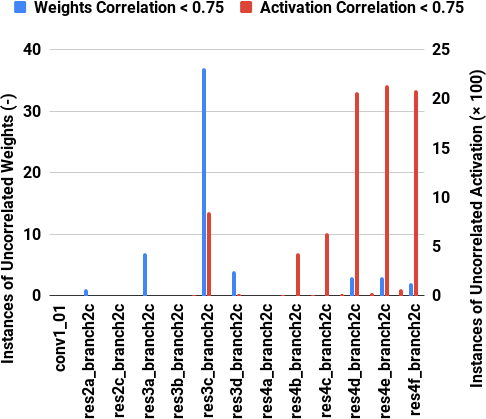}
    \caption{The comparison of the original and fine-tuned \textit{ResNet50}. Blue bars refer to the total number of uncorrelated kernels' weights (\ie $corr \leq 0.75$) between networks. Red bars refer to the total number of images in validation-set of \textit{ImageNet} with uncorrelated kernels' activation (\ie $corr \leq 0.75$) between both networks.}
    \label{fig:r_w_a}
\end{figure}

Another approach to compare a pair of twin networks is to first feed each with an image, in order to compute the activation map of all kernels, and then compare those correspondingly. We conducted this experiment for all layers of the original and fine-tuned \textit{ResNet50}, inputting them with all images in validation-set of \textit{ImageNet}. Representing each layer through its average across all 50K images exhibits an overall tendency between lower and higher layers, similar to curves of Figure \ref{fig:weights}. Alternatively, each image can be examined conditioned to a nominal threshold. We observed interesting correspondences between this analysis and the one for kernels' weights, refer to Figure \ref{fig:r_w_a}. Most ``\textit{branch\_2c}" convolutional layers stand out as completely different from other layers (note red bars in this figure).

Kernels of ``\textit{branch\_2c}" perform convolution only along the dimension of feature maps without integration over a spatial vicinity (\ie kernels of size $1 \times 1$). This construction is a reminiscence of hypercolumns in biological vision~\cite{hubel1963shape}. Perhaps, these $1 \times 1$ kernels that expand or contract dimensionality of feature maps allow the network to learn an account of image contrast. Therefore, it can be contemplated that future investigations concentrated on these kernels could shed more light on the exact operations that lead to contrast invariance, at least in case of \textit{ResNet50}.

\subsubsection{Successful versus failure comparison}

Ultimately, any attempt to describe the mechanism of contrast invariance in DNNs (or any other feature as a matter of fact) should be able to explain differences at the level of individual images as well. We inspected all the discovered patterns in addition to the one proposed in \cite{akbarinia2018contrast} with this criteria. For instance, whether kernels of ``\textit{branch\_2c}" in \textit{ResNet50} exhibit distinct activation maps for successful and failure trials. We did not spot any systematic differences between the two conditions: in other words the shape of red bars in Figure \ref{fig:r_w_a} would be almost identical regardless of network output (refer to supplementary materials).

\section{Conclusion}

In this work we enquired into the function of image contrast in deep neural networks (DNN). We argued that contrast is a pillar of a visual system as manifested in evolution of the biological vision \cite{mante2005independence}. We further reasoned that invariance to this feature is a necessary asset for machine vision, evident in case of autonomous cars \cite{tian2018deeptest}. We addressed two research questions: (i) \textbf{how} to accomplish a \textit{contrast-invariant} model, and (ii) \textbf{what} mechanism allows a DNN to possess this feature. We approached these questions by conducting experiments on \textit{ImageNet}: a large visual data set of diverse objects. We thoroughly studied thirteen prominent networks in the literature by fine-tuning their weights with \textit{contrast-augmentation}, as well as training new instances of each architecture under completely controlled conditions.

The results of our experiments report that state-of-the-art object classification networks fail to retain their peak performance for images of poor contrast, some even suffer at higher levels as much as 50\% (refer to Figure \ref{fig:prominents} and Table~\ref{tab:all}). We demonstrated that a simple fine-tuning procedure of \textit{contrast-augmentation} offers a robust solution to the \textbf{how} question: by allowing a model to adequately adjust its parameters to perform almost perfectly at extreme low-contrast. Training new instances from scratch supported these findings by exhibiting that exposure to multiple levels of contrast is indeed the key factor (see Figure~\ref{fig:trained}).

We tackled the \textbf{what} question by comparing weights and activation maps of every layer and its constituent kernels between an original network and its fine-tuned offspring. These networks are over 99.9\% correlated, however one is invariant to image contrast while another heavily impaired by it. We observed a general tendency that the largest difference appears to occur at the first few layers. Other pronounced patterns emerged uniquely for each architecture, however none explained the difference between successful and failure trials. A more profound study of those would allow future works to answer the what question.

Deciphering the exact mechanisms of a system constitutes of millions of parameters is complex due to the convoluted nature of its inner operations. DNN is a hyper-dimensional model, however for simplicity it can be visualised as a prism of multiple areas (\ie a block of repetitive layers). Each of these areas could be imagined as a cuboid decomposed into layers of different nature. Each layer is a volumetric shape sliced to kernels that are the smallest comprehensible dimension of a network due to their three dimensional nature. In this article (similar to previous works \cite{zeiler2014visualizing,mahendran2015understanding,bau2017network}) we limited our analysis to the kernel and layer dimension. Given hierarchical depth is demonstrated to be of significant importance both in artificial \cite{simonyan2014very,szegedy2015going,he2015delving} and biological vision \cite{riesenhuber1999hierarchical,hochstein2002view,kruger2013deep}. Therefore, future works should focus more on the analysis of hyper dimensions (\ie the \textbf{depth} of a neural network, across layers and areas).

\subsubsection*{Acknowledgements}

This project was funded by the Deutsche Forschungsgemeinschaft SFB/TRR 135.

{\small
\bibliographystyle{ieee}
\bibliography{egbib}

\begin{thebibliography}{10}\itemsep=-1pt

\bibitem{agrawal2014analyzing}
P.~Agrawal, R.~Girshick, and J.~Malik.
\newblock Analyzing the performance of multilayer neural networks for object
  recognition.
\newblock In {\em European conference on computer vision}, pages 329--344.
  Springer, 2014.

\bibitem{akbarinia2018contrast}
A.~Akbarinia and K.~R. Gegenfurtner.
\newblock How is contrast encoded in deep neural networks?
\newblock {\em arXiv preprint arXiv:1809.01438}, 2018.

\bibitem{akbarinia2017feedback}
A.~Akbarinia and C.~A. Parraga.
\newblock Feedback and surround modulated boundary detection.
\newblock {\em International Journal of Computer Vision}, pages 1--14, 2017.

\bibitem{akbarinia2018colour}
A.~Akbarinia and C.~A. Parraga.
\newblock Colour constancy beyond the classical receptive field.
\newblock {\em IEEE transactions on pattern analysis and machine intelligence},
  40(9):2081--2094, 2018.

\bibitem{bau2017network}
D.~Bau, B.~Zhou, A.~Khosla, A.~Oliva, and A.~Torralba.
\newblock Network dissection: Quantifying interpretability of deep visual
  representations.
\newblock {\em arXiv preprint arXiv:1704.05796}, 2017.

\bibitem{bengio2013representation}
Y.~Bengio, A.~Courville, and P.~Vincent.
\newblock Representation learning: A review and new perspectives.
\newblock {\em IEEE transactions on pattern analysis and machine intelligence},
  35(8):1798--1828, 2013.

\bibitem{chollet2017xception}
F.~Chollet.
\newblock Xception: Deep learning with depthwise separable convolutions.
\newblock {\em arXiv preprint}, pages 1610--02357, 2017.

\bibitem{dodge2016understanding}
S.~Dodge and L.~Karam.
\newblock Understanding how image quality affects deep neural networks.
\newblock In {\em Quality of Multimedia Experience (QoMEX), 2016 Eighth
  International Conference on}, pages 1--6. IEEE, 2016.

\bibitem{frazor2006local}
R.~A. Frazor and W.~S. Geisler.
\newblock Local luminance and contrast in natural images.
\newblock {\em Vision research}, 46(10):1585--1598, 2006.

\bibitem{geirhos2017comparing}
R.~Geirhos, D.~H. Janssen, H.~H. Sch{\"u}tt, J.~Rauber, M.~Bethge, and F.~A.
  Wichmann.
\newblock Comparing deep neural networks against humans: object recognition
  when the signal gets weaker.
\newblock {\em arXiv preprint arXiv:1706.06969}, 2017.

\bibitem{gonzalez2018semantic}
A.~Gonzalez-Garcia, D.~Modolo, and V.~Ferrari.
\newblock Do semantic parts emerge in convolutional neural networks?
\newblock {\em International Journal of Computer Vision}, 126(5):476--494,
  2018.

\bibitem{he2015delving}
K.~He, X.~Zhang, S.~Ren, and J.~Sun.
\newblock Delving deep into rectifiers: Surpassing human-level performance on
  imagenet classification.
\newblock In {\em Proceedings of the IEEE international conference on computer
  vision}, pages 1026--1034, 2015.

\bibitem{he2016deep}
K.~He, X.~Zhang, S.~Ren, and J.~Sun.
\newblock Deep residual learning for image recognition.
\newblock In {\em Proceedings of the IEEE conference on computer vision and
  pattern recognition}, pages 770--778, 2016.

\bibitem{hochstein2002view}
S.~Hochstein and M.~Ahissar.
\newblock View from the top: Hierarchies and reverse hierarchies in the visual
  system.
\newblock {\em Neuron}, 36(5):791--804, 2002.

\bibitem{howard2017mobilenets}
A.~G. Howard, M.~Zhu, B.~Chen, D.~Kalenichenko, W.~Wang, T.~Weyand,
  M.~Andreetto, and H.~Adam.
\newblock Mobilenets: Efficient convolutional neural networks for mobile vision
  applications.
\newblock {\em arXiv preprint arXiv:1704.04861}, 2017.

\bibitem{huang2017densely}
G.~Huang, Z.~Liu, L.~Van Der~Maaten, and K.~Q. Weinberger.
\newblock Densely connected convolutional networks.
\newblock In {\em CVPR}, volume~1, page~3, 2017.

\bibitem{hubel1963shape}
D.~H. Hubel and T.~Wiesel.
\newblock Shape and arrangement of columns in cat's striate cortex.
\newblock {\em The Journal of physiology}, 165(3):559--568, 1963.

\bibitem{hubel1962receptive}
D.~H. Hubel and T.~N. Wiesel.
\newblock Receptive fields, binocular interaction and functional architecture
  in the cat's visual cortex.
\newblock {\em The Journal of physiology}, 160(1):106--154, 1962.

\bibitem{iandola2016squeezenet}
F.~N. Iandola, S.~Han, M.~W. Moskewicz, K.~Ashraf, W.~J. Dally, and K.~Keutzer.
\newblock Squeezenet: Alexnet-level accuracy with 50x fewer parameters and< 0.5
  mb model size.
\newblock {\em arXiv preprint arXiv:1602.07360}, 2016.

\bibitem{itti2001computational}
L.~Itti and C.~Koch.
\newblock Computational modelling of visual attention.
\newblock {\em Nature reviews neuroscience}, 2(3):194, 2001.

\bibitem{kandel2000principles}
E.~R. Kandel, J.~H. Schwartz, T.~M. Jessell, D.~of~Biochemistry, M.~B.~T.
  Jessell, S.~Siegelbaum, and A.~Hudspeth.
\newblock {\em Principles of neural science}, volume~4.
\newblock McGraw-hill New York, 2000.

\bibitem{kingma2014adam}
D.~P. Kingma and J.~Ba.
\newblock Adam: A method for stochastic optimization.
\newblock {\em arXiv preprint arXiv:1412.6980}, 2014.

\bibitem{kokkinos2017ubernet}
I.~Kokkinos.
\newblock Ubernet: Training a universal convolutional neural network for low-,
  mid-, and high-level vision using diverse datasets and limited memory.
\newblock In {\em CVPR}, volume~2, page~8, 2017.

\bibitem{krizhevsky2012imagenet}
A.~Krizhevsky, I.~Sutskever, and G.~E. Hinton.
\newblock Imagenet classification with deep convolutional neural networks.
\newblock In {\em Advances in neural information processing systems}, pages
  1097--1105, 2012.

\bibitem{kruger2013deep}
N.~Kruger, P.~Janssen, S.~Kalkan, M.~Lappe, A.~Leonardis, J.~Piater, A.~J.
  Rodriguez-Sanchez, and L.~Wiskott.
\newblock Deep hierarchies in the primate visual cortex: What can we learn for
  computer vision?
\newblock {\em IEEE transactions on pattern analysis and machine intelligence},
  35(8):1847--1871, 2013.

\bibitem{lecun2015deep}
Y.~LeCun, Y.~Bengio, and G.~Hinton.
\newblock Deep learning.
\newblock {\em nature}, 521(7553):436, 2015.

\bibitem{li2017learning}
Z.~Li and D.~Hoiem.
\newblock Learning without forgetting.
\newblock {\em IEEE Transactions on Pattern Analysis and Machine Intelligence},
  2017.

\bibitem{mahendran2015understanding}
A.~Mahendran and A.~Vedaldi.
\newblock Understanding deep image representations by inverting them.
\newblock In {\em Proceedings of the IEEE conference on computer vision and
  pattern recognition}, pages 5188--5196, 2015.

\bibitem{mante2005independence}
V.~Mante, R.~A. Frazor, V.~Bonin, W.~S. Geisler, and M.~Carandini.
\newblock Independence of luminance and contrast in natural scenes and in the
  early visual system.
\newblock {\em Nature neuroscience}, 8(12):1690, 2005.

\bibitem{michelson1995studies}
A.~A. Michelson.
\newblock {\em Studies in optics}.
\newblock Courier Corporation, 1995.

\bibitem{perazzi2012saliency}
F.~Perazzi, P.~Kr{\"a}henb{\"u}hl, Y.~Pritch, and A.~Hornung.
\newblock Saliency filters: Contrast based filtering for salient region
  detection.
\newblock In {\em Computer Vision and Pattern Recognition (CVPR), 2012 IEEE
  Conference on}, pages 733--740. IEEE, 2012.

\bibitem{purves2011we}
D.~Purves and R.~B. Lotto.
\newblock {\em Why we see what we do redux: A wholly empirical theory of
  vision.}
\newblock Sinauer Associates, 2011.

\bibitem{rastegari2016xnor}
M.~Rastegari, V.~Ordonez, J.~Redmon, and A.~Farhadi.
\newblock Xnor-net: Imagenet classification using binary convolutional neural
  networks.
\newblock In {\em European Conference on Computer Vision}, pages 525--542.
  Springer, 2016.

\bibitem{riesenhuber1999hierarchical}
M.~Riesenhuber and T.~Poggio.
\newblock Hierarchical models of object recognition in cortex.
\newblock {\em Nature neuroscience}, 2(11):1019, 1999.

\bibitem{sandler2018mobilenetv2}
M.~Sandler, A.~Howard, M.~Zhu, A.~Zhmoginov, and L.-C. Chen.
\newblock Mobilenetv2: Inverted residuals and linear bottlenecks.
\newblock In {\em Proceedings of the IEEE Conference on Computer Vision and
  Pattern Recognition}, pages 4510--4520, 2018.

\bibitem{schlyter2009radiometry}
P.~Schlyter.
\newblock Radiometry and photometry in astronomy.
\newblock {\em Available: stjarnhimlen. se/comp/radfaq. html}, 2009.

\bibitem{simonyan2014very}
K.~Simonyan and A.~Zisserman.
\newblock Very deep convolutional networks for large-scale image recognition.
\newblock {\em arXiv preprint arXiv:1409.1556}, 2014.

\bibitem{szegedy2017inception}
C.~Szegedy, S.~Ioffe, V.~Vanhoucke, and A.~A. Alemi.
\newblock Inception-v4, inception-resnet and the impact of residual connections
  on learning.
\newblock In {\em AAAI}, volume~4, page~12, 2017.

\bibitem{szegedy2015going}
C.~Szegedy, W.~Liu, Y.~Jia, P.~Sermanet, S.~Reed, D.~Anguelov, D.~Erhan,
  V.~Vanhoucke, and A.~Rabinovich.
\newblock Going deeper with convolutions.
\newblock In {\em Proceedings of the IEEE conference on computer vision and
  pattern recognition}, pages 1--9, 2015.

\bibitem{szegedy2016rethinking}
C.~Szegedy, V.~Vanhoucke, S.~Ioffe, J.~Shlens, and Z.~Wojna.
\newblock Rethinking the inception architecture for computer vision.
\newblock In {\em Proceedings of the IEEE conference on computer vision and
  pattern recognition}, pages 2818--2826, 2016.

\bibitem{tian2018deeptest}
Y.~Tian, K.~Pei, S.~Jana, and B.~Ray.
\newblock Deeptest: Automated testing of deep-neural-network-driven autonomous
  cars.
\newblock In {\em Proceedings of the 40th International Conference on Software
  Engineering}, pages 303--314. ACM, 2018.

\bibitem{zeiler2014visualizing}
M.~D. Zeiler and R.~Fergus.
\newblock Visualizing and understanding convolutional networks.
\newblock In {\em European conference on computer vision}, pages 818--833.
  Springer, 2014.

\bibitem{zoph2017learning}
B.~Zoph, V.~Vasudevan, J.~Shlens, and Q.~V. Le.
\newblock Learning transferable architectures for scalable image recognition.
\newblock {\em arXiv preprint arXiv:1707.07012}, 2(6), 2017.

\end{thebibliography}
}

\appendix 

\section{Pretrained networks}

\subsection{Contrast reduction}

Classification accuracy of various networks under seven levels of image contrast -- $c \in \left\{ 1, 5, 15, 30, 50, 75, 100 \right\} \%$ -- are reported in Figure \ref{fig:prominents}. Refer to Section 3.1 and 4.1 of the principal manuscript for relevant discussion.

\begin{figure*}[ht]
    \centering
    \begin{tabular}{cc}
         \includegraphics[width=0.45\textwidth]{prominents1.png} & \includegraphics[width=0.45\textwidth]{prominents2.png} \\
         \includegraphics[width=0.45\textwidth]{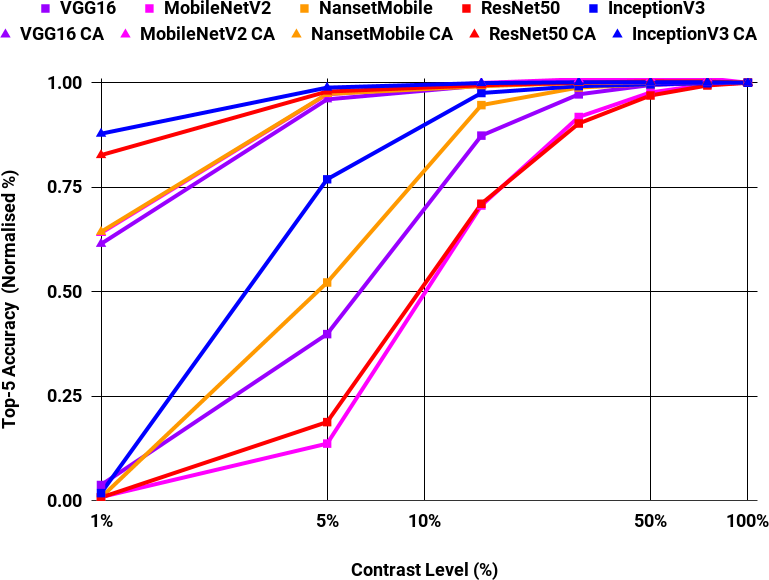} & \includegraphics[width=0.45\textwidth]{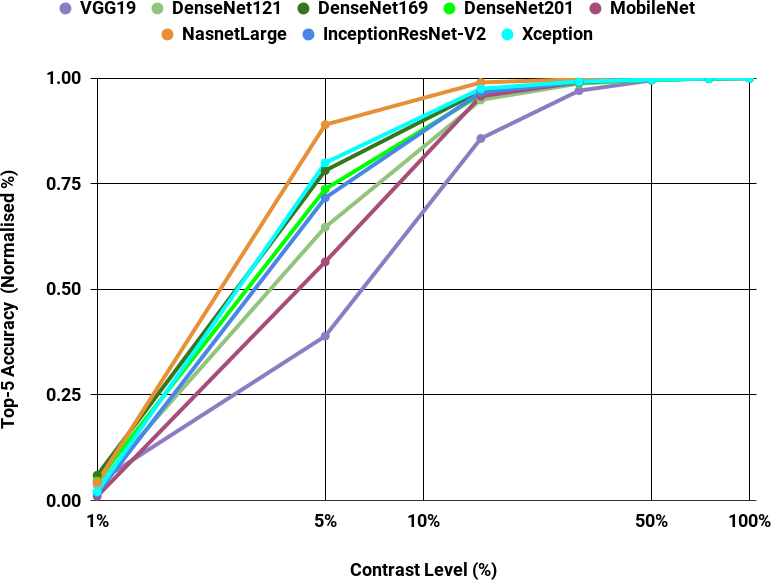}         
    \end{tabular}
    \caption{The classification accuracy of various networks on the validation-set of \textit{ImageNet} data set. The first row corresponds to top-1 and the second row corresponds to top-5. The curves are \textbf{normalised} to perfect accuracy on 100\% level of contrast. On the left panel: those curves with a \textbf{triangle} shape have gone through \textit{contrast-augmented} fine-tuning, initialised with the weights of the \textbf{square} shaped curves. On the right panel: all the rest of the pretrained models obtained from Keras platform.}
    \label{fig:prominents}
\end{figure*}

\subsection{Illumination manipulation}

\begin{figure*}[ht]
    \centering
    \begin{tabular}{cc}
         \includegraphics[width=0.45\textwidth]{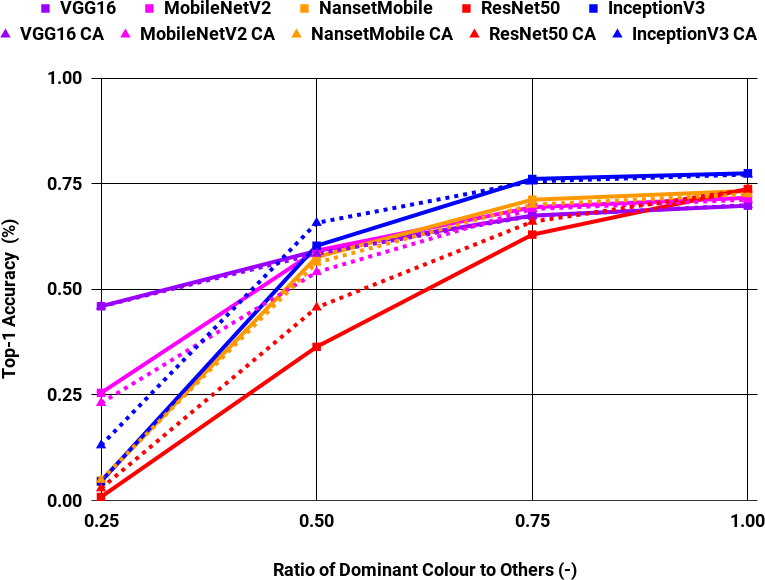} & \includegraphics[width=0.45\textwidth]{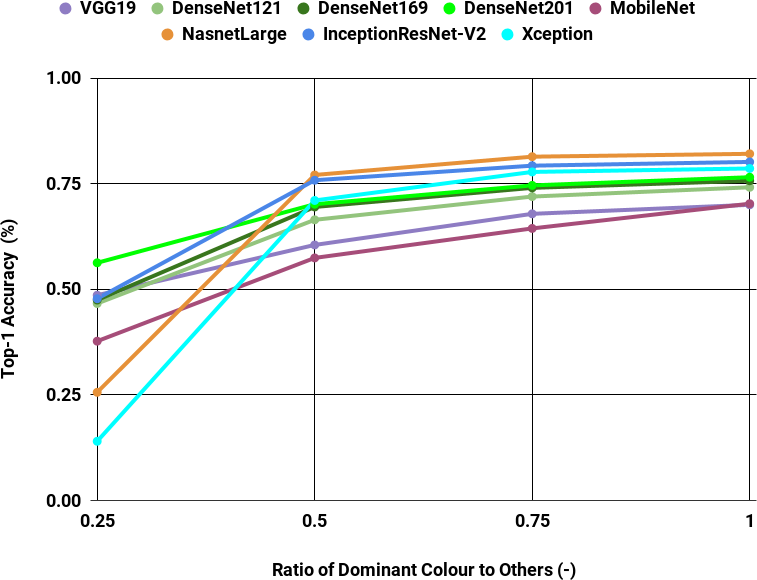} \\
         \includegraphics[width=0.45\textwidth]{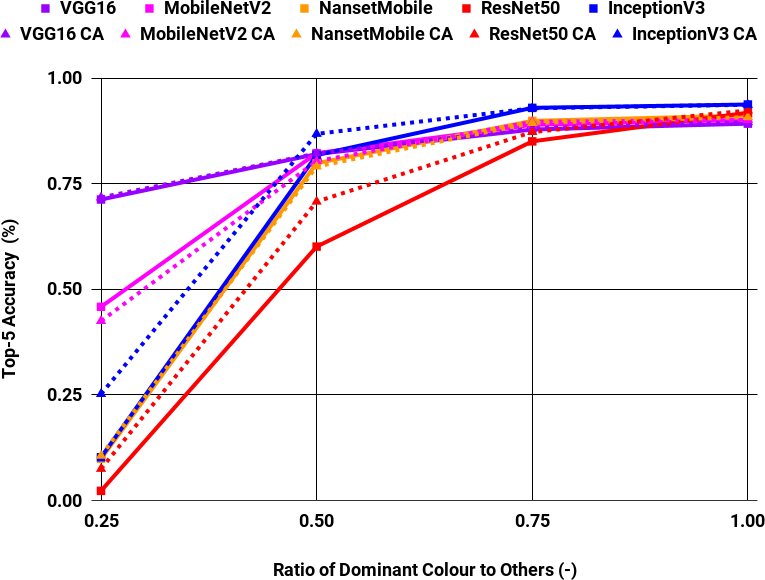} & \includegraphics[width=0.45\textwidth]{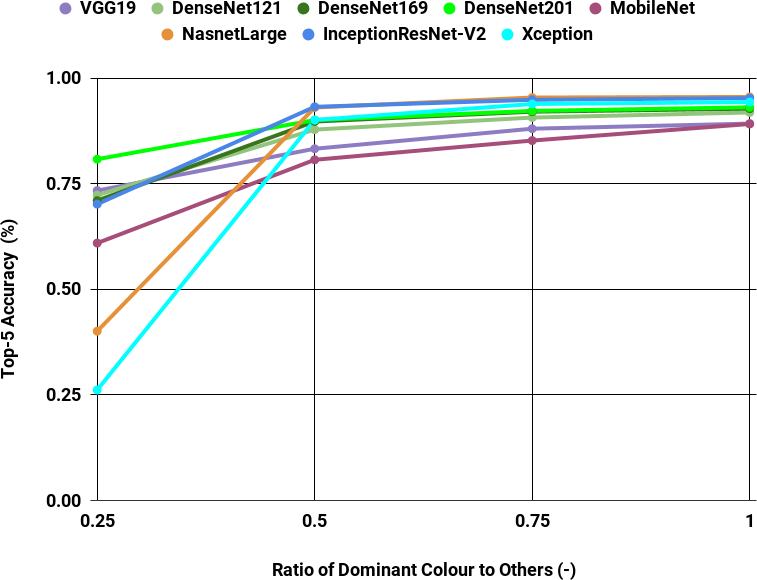}
    \end{tabular}
    \caption{The classification accuracy of various networks on the validation-set of \textit{ImageNet} data set. The first row corresponds to top-1 and the second row corresponds to top-5. On the left panel: those curves with a \textbf{triangle} shape and dotted pattern have gone through \textit{contrast-augmented} fine-tuning, initialised with the weights of the \textbf{square} shaped curves. On the right panel: all the rest of the pretrained models obtained from Keras platform.}
    \label{fig:constancy}
\end{figure*}

Classification accuracy of various networks under three conditions of illumination manipulation are reported in Figure \ref{fig:constancy}. The reported results are average of three conditions: \ie reddish, greenish, or bluish (obtained by keeping one colour channel constant and multiplying the other two channels with a constant of 0.25, 0.50, and 0.75). There is no significant difference between \textit{VGG16}, \textit{MobileNetV2}, \textit{NasnetMobile} and their corresponding contrast-augmented fine-tuned offspring. However, the fine-tuned \textit{InceptionV3} and \textit{ResNet50} score significantly better than their original networks.

Although not a topic of this article, it is worth highlighting that certain networks (\textit{DenseNet} and \textit{VGG} family) perform substantially better in this task in comparison to the others. This should be investigated in future studies.

\subsection{Gamma correction}

\begin{figure*}[ht]
    \centering
    \begin{tabular}{cc}
         \includegraphics[width=0.45\textwidth]{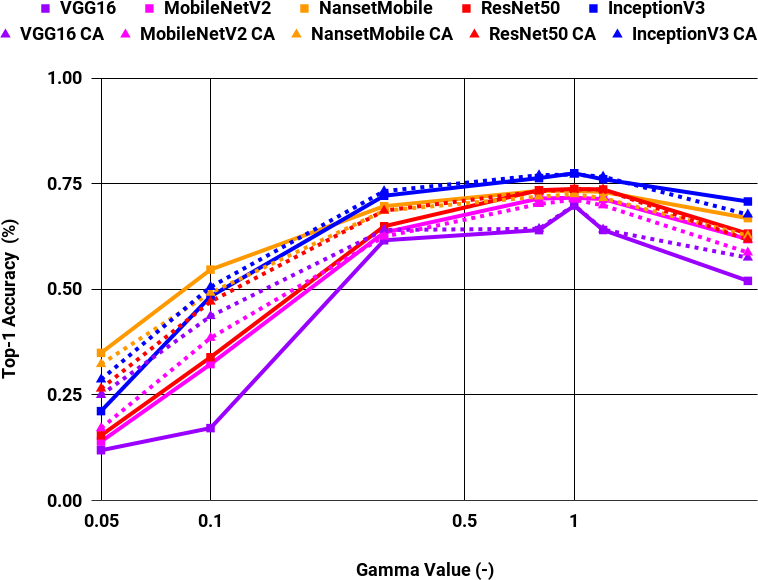} & \includegraphics[width=0.45\textwidth]{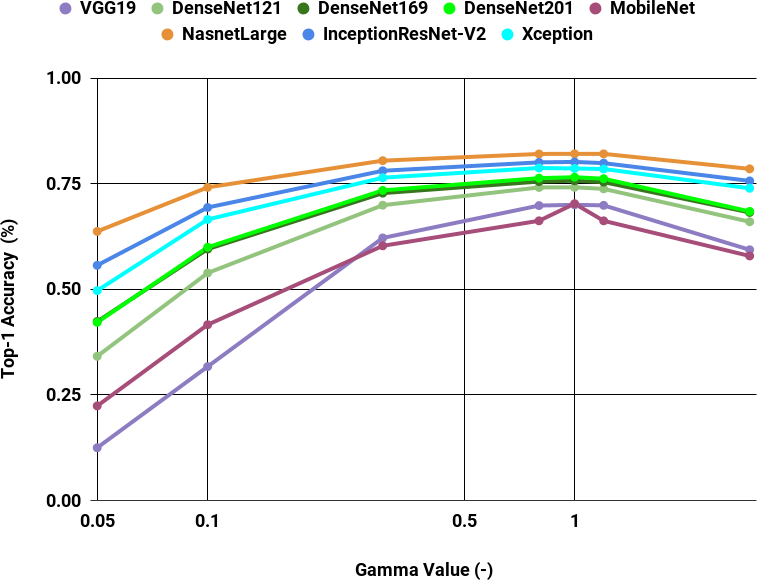} \\
         \includegraphics[width=0.45\textwidth]{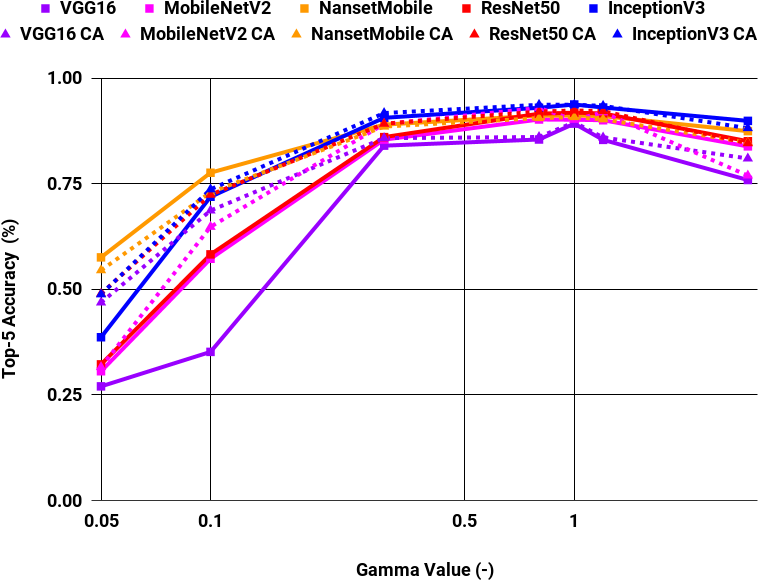} & \includegraphics[width=0.45\textwidth]{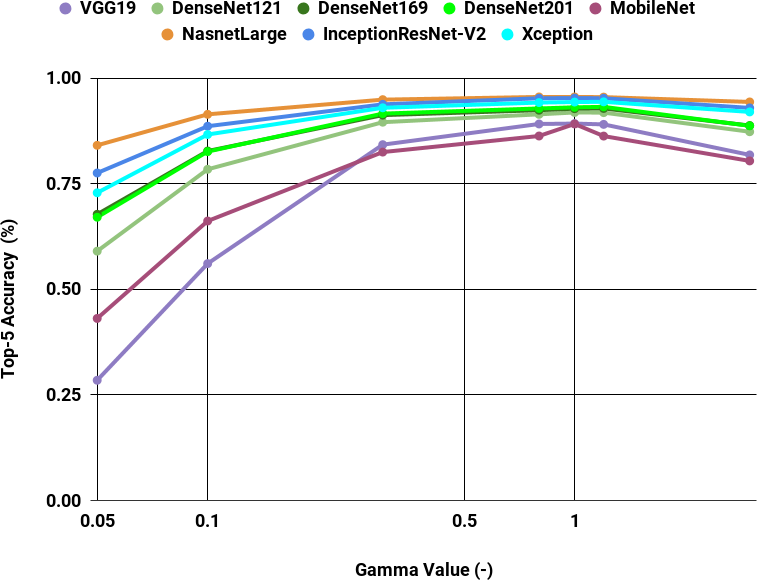}         
    \end{tabular}
    \caption{The classification accuracy of various networks on the validation-set of \textit{ImageNet} data set. The first row corresponds to top-1 and the second row corresponds to top-5. On the left panel: those curves with a \textbf{triangle} shape and dotted pattern have gone through \textit{contrast-augmented} fine-tuning, initialised with the weights of the \textbf{square} shaped curves. On the right panel: all the rest of the pretrained models obtained from Keras platform.}
    \label{fig:gamma}
\end{figure*}

Classification accuracy of various networks under seven Gamma correction -- $\gamma \in \left\{ 0.05,0.1, 0.3, 0.8, 1, 1.2, 3 \right\}$ -- are reported in Figure \ref{fig:gamma}. All contrast-augmented fine-tuned networks perform considerably better in $\gamma < 1$. This is more pronounced for \textit{VGG16} and \textit{ResNet50}. For $\gamma > 1$ there is no clear pattern between original and fine-tuned offspring.

Although not a topic of this article, it is worth highlighting that certain networks (\textit{NasnetLarge}) perform substantially better in this task in comparison to the others. This should be investigated in future studies.

\subsection{Gaussian blurring}

\begin{figure*}[ht]
    \centering
    \begin{tabular}{cc}
         \includegraphics[width=0.45\textwidth]{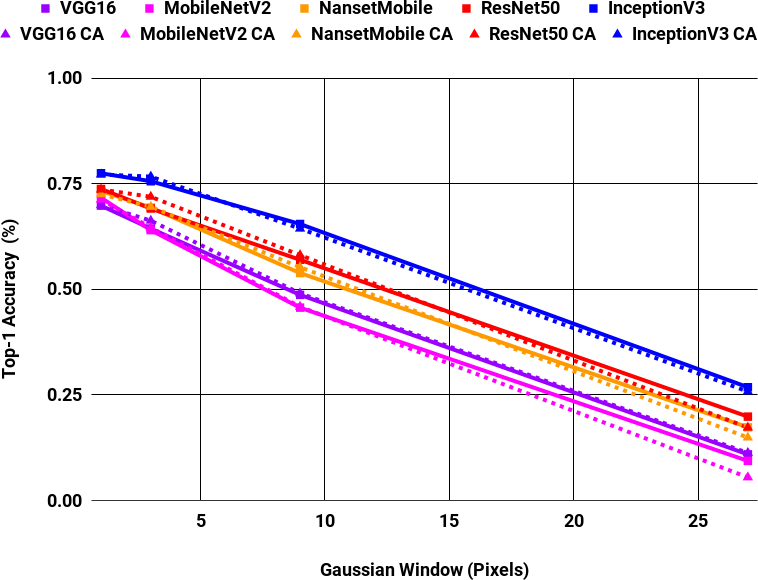} & \includegraphics[width=0.45\textwidth]{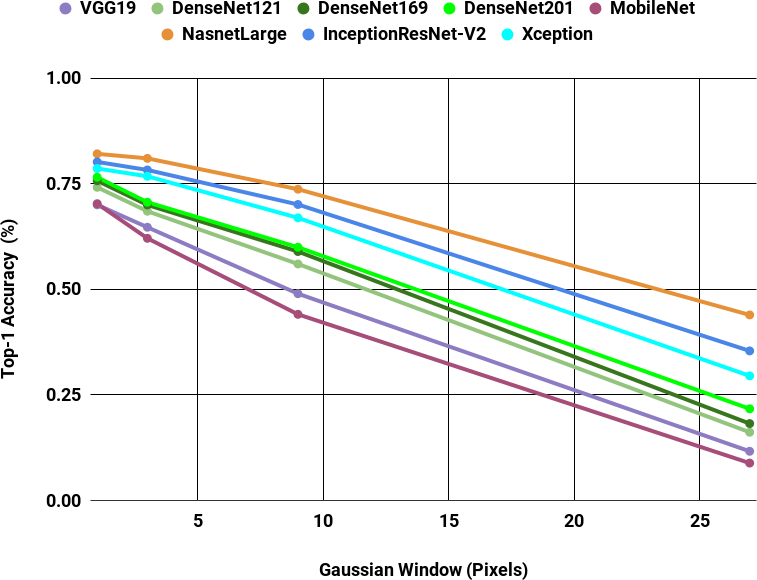} \\
         \includegraphics[width=0.45\textwidth]{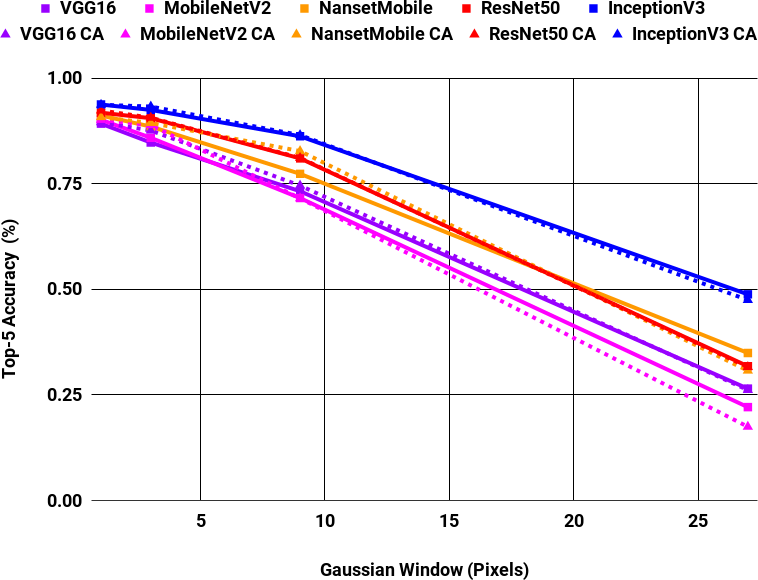} & \includegraphics[width=0.45\textwidth]{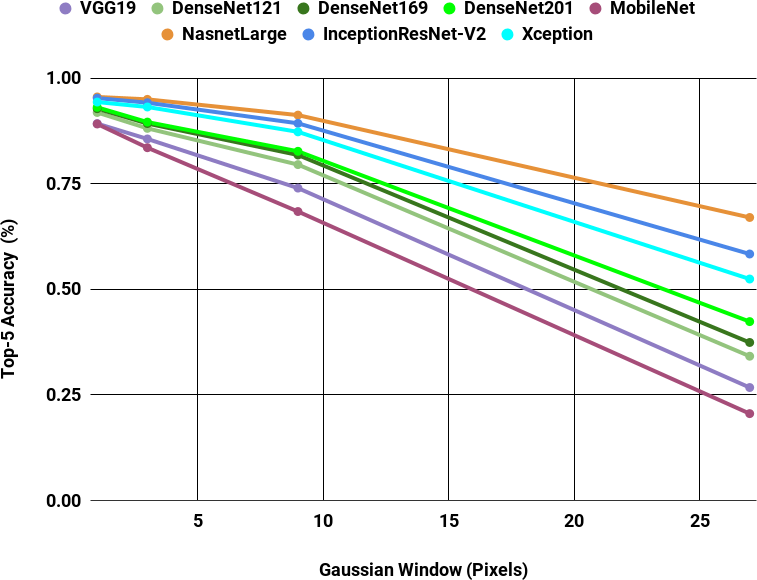}         
    \end{tabular}
    \caption{The classification accuracy of various networks on the validation-set of \textit{ImageNet} data set. The first row corresponds to top-1 and the second row corresponds to top-5. On the left panel: those curves with a \textbf{triangle} shape and dotted pattern have gone through \textit{contrast-augmented} fine-tuning, initialised with the weights of the \textbf{square} shaped curves. On the right panel: all the rest of the pretrained models obtained from Keras platform.}
    \label{fig:gaussian}
\end{figure*}

Classification accuracy of various networks under three levels of Gaussian blurring -- convolutional window of side 3, 9 and 27 pixels -- are reported in Figure \ref{fig:gaussian}. There is no significant difference between original and fine-tuned offspring.

Although not a topic of this article, it is worth highlighting that certain networks (\textit{NasnetLarge}) perform substantially better in this task in comparison to the others. This should be investigated in future studies.

\subsection{Uniform noise}

\begin{figure*}[ht]
    \centering
    \begin{tabular}{cc}
         \includegraphics[width=0.45\textwidth]{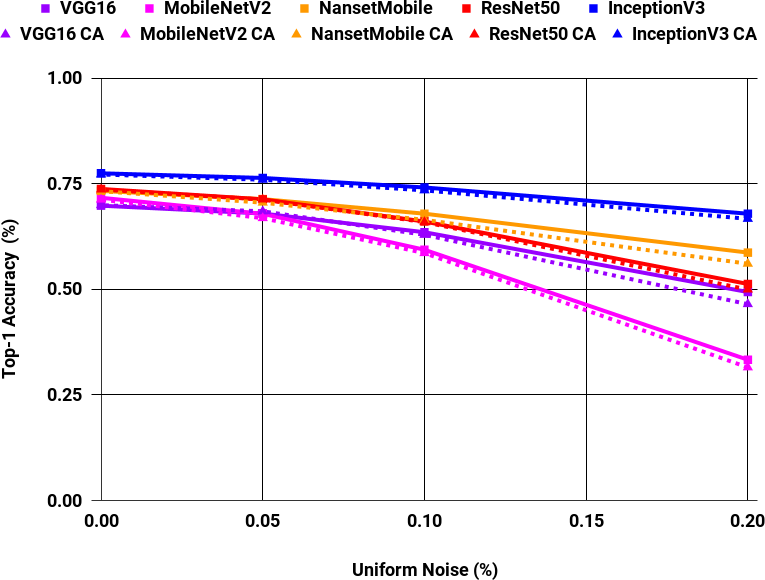} & \includegraphics[width=0.45\textwidth]{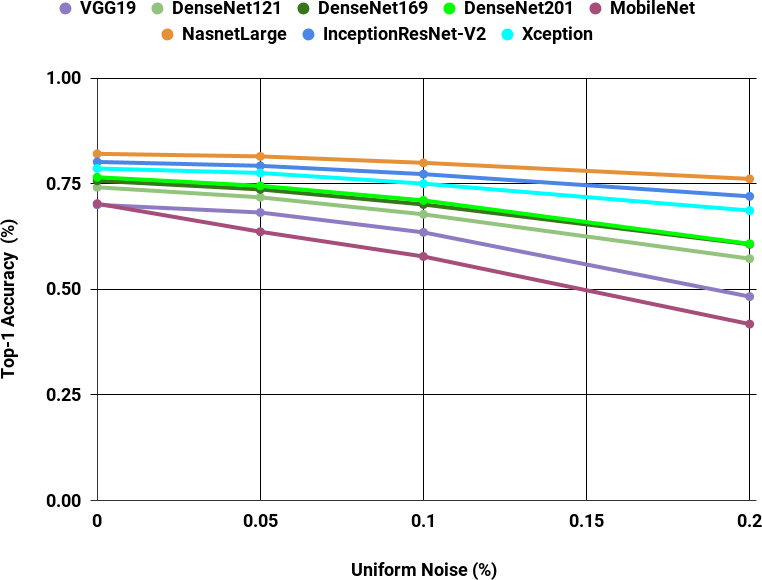} \\
         \includegraphics[width=0.45\textwidth]{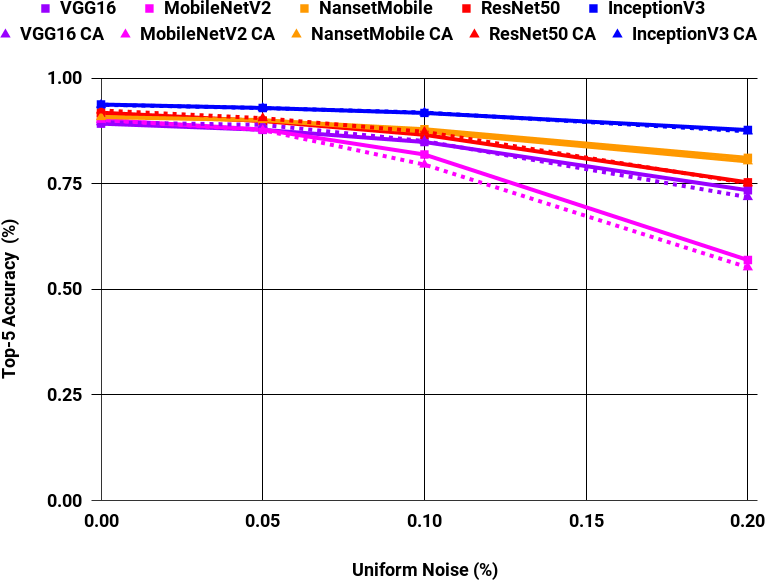} & \includegraphics[width=0.45\textwidth]{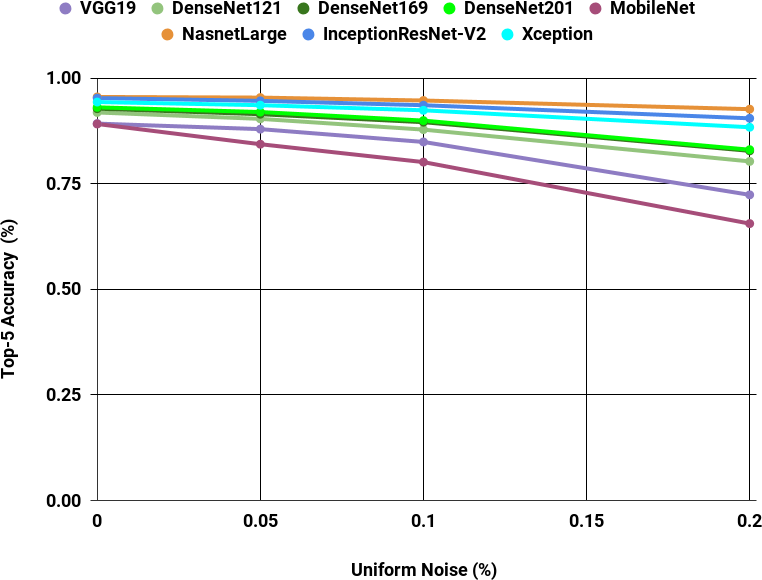}
    \end{tabular}
    \caption{The classification accuracy of various networks on the validation-set of \textit{ImageNet} data set. The first row corresponds to top-1 and the second row corresponds to top-5. On the left panel: those curves with a \textbf{triangle} shape and dotted pattern have gone through \textit{contrast-augmented} fine-tuning, initialised with the weights of the \textbf{square} shaped curves. On the right panel: all the rest of the pretrained models obtained from Keras platform.}
    \label{fig:uniform}
\end{figure*}

Classification accuracy of various networks under three levels of uniform noise -- 5, 10, and 20\% noise -- are reported in Figure \ref{fig:uniform}. There is no significant difference between original and fine-tuned offspring.

Although not a topic of this article, it is worth highlighting that certain networks (\textit{NasnetLarge}, \textit{InceptionV3}, \textit{Xception}, and \textit{InceptionResNetV2}) perform substantially better in this task in comparison to the others. This should be investigated in future studies.

\subsection{Salt \& Pepper Noise}

\begin{figure*}[ht]
    \centering
    \begin{tabular}{cc}
         \includegraphics[width=0.45\textwidth]{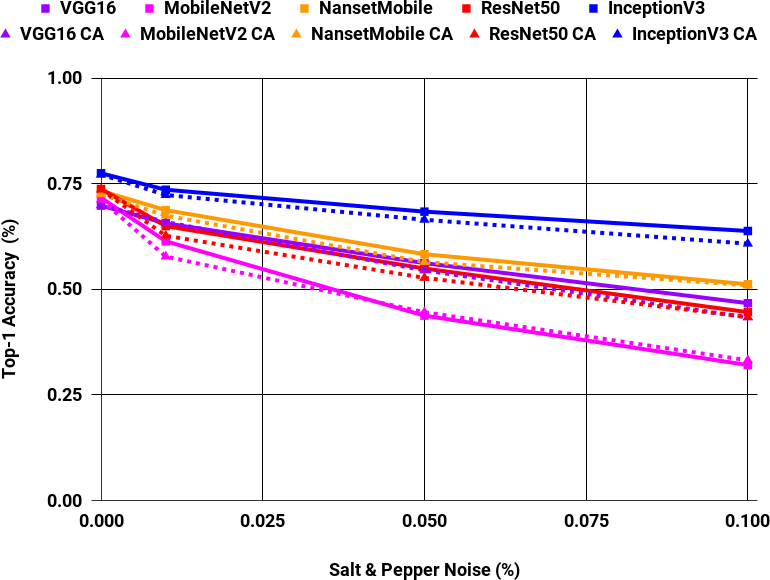} & \includegraphics[width=0.45\textwidth]{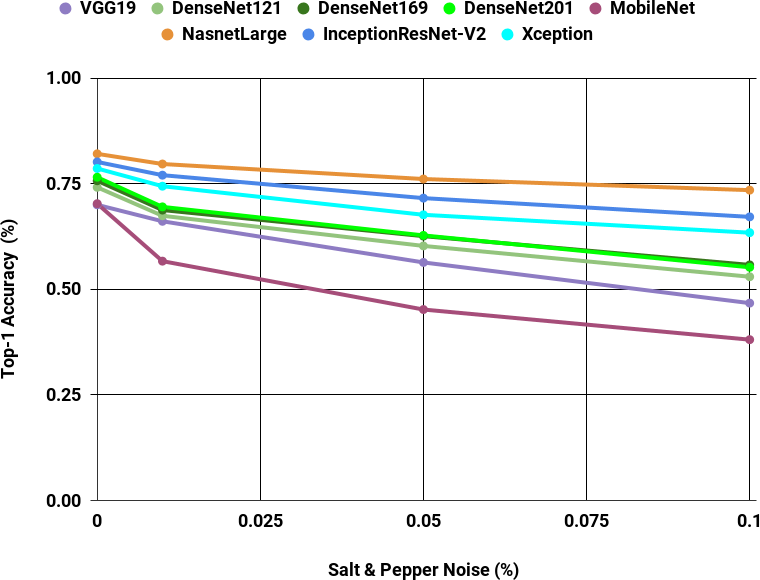} \\
         \includegraphics[width=0.45\textwidth]{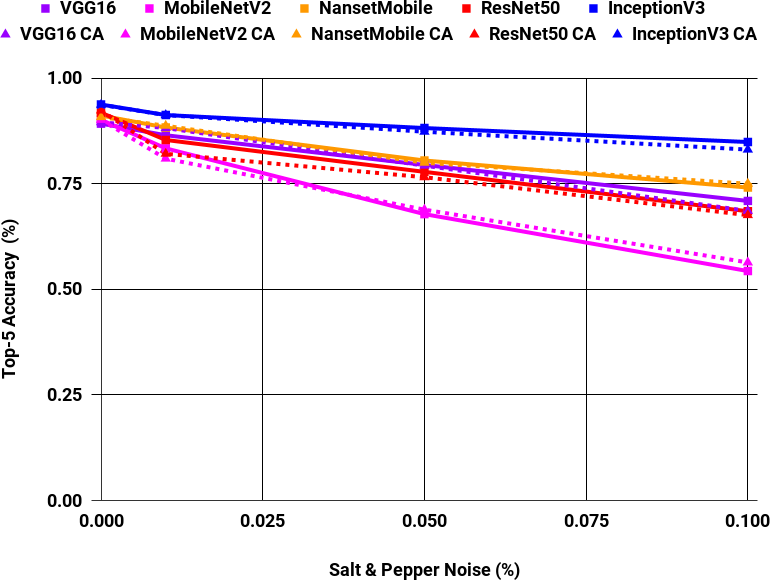} & \includegraphics[width=0.45\textwidth]{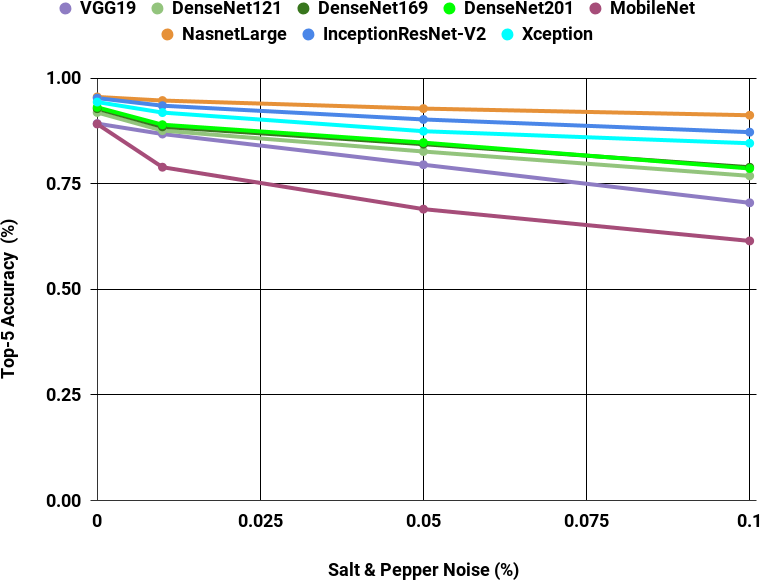}   
    \end{tabular}
    \caption{The classification accuracy of various networks on the validation-set of \textit{ImageNet} data set. The first row corresponds to top-1 and the second row corresponds to top-5. On the left panel: those curves with a \textbf{triangle} shape and dotted pattern have gone through \textit{contrast-augmented} fine-tuning, initialised with the weights of the \textbf{square} shaped curves. On the right panel: all the rest of the pretrained models obtained from Keras platform.}
    \label{fig:sandp}
\end{figure*}

Classification accuracy of various networks under three levels of salt \& pepper noise -- 1, 5, and 10\% noise -- are reported in Figure \ref{fig:sandp}. There is no significant difference between original and fine-tuned offspring.

Although not a topic of this article, it is worth highlighting that certain networks (\textit{NasnetLarge}) perform substantially better in this task in comparison to the others. This should be investigated in future studies.

\subsection{Amount of required training}

\begin{figure*}[ht]
    \centering
    \begin{tabular}{cc}
        \includegraphics[width=0.45\textwidth]{resnet_epochs.png} & \includegraphics[width=0.45\textwidth]{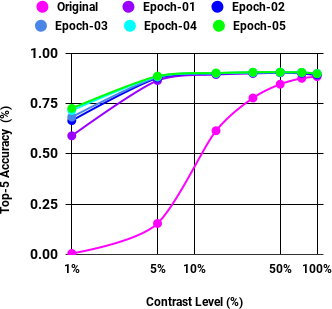}
    \end{tabular}
    \caption{The top-1 and top-5 classification accuracy of fine-tuned \textit{ResNet50} in various epochs on the validation-set of \textit{ImageNet} data set.}
    \label{fig:evolution}
\end{figure*}

\section{Training networks from scratch}
\subsection{Contrast reduction}
Classification accuracy of various networks trained in controlled environment under seven levels of image contrast -- $c \in \left\{ 1, 5, 15, 30, 50, 75, 100 \right\} \%$ -- are reported in Figure \ref{fig:trained}. Refer to Section 3.2 of the principal manuscript for relevant discussion.

\begin{figure*}[ht]
    \centering
    \begin{tabular}{cc}
        \includegraphics[width=0.45\textwidth]{trrained.png} & \includegraphics[width=0.45\textwidth]{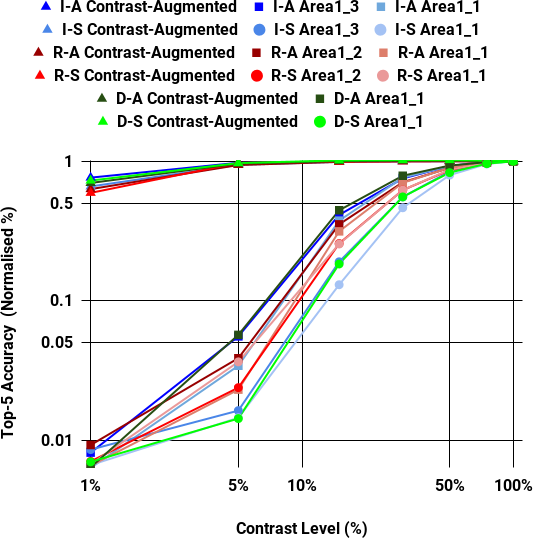}
    \end{tabular}
    \caption{The top-1 and top-5 classification accuracy of various networks on the validation-set of \textit{ImageNet}. The left panel corresponds to top-1 and the right panel corresponds to top-5 The curves are \textbf{normalised} to perfect accuracy on 100\% level of contrast. Each legend starts with an \textbf{abbreviation}, the first corresponds to network (\textit{I: InceptionV3} -- \textbf{Blue}, \textit{R: ResNet50} -- \textbf{Red}, \textit{D: DenseNet201} -- \textbf{Green}) and the second to optimiser (\textit{A: Adam} -- \textbf{Square}, \textit{S: SGD} -- \textbf{Circle}). Those with a \textbf{triangle} shape have gone through a \textit{contrast-augmented} training procedure.}
    \label{fig:trained}
\end{figure*}

\section{The what question}
\subsection{Weights comparison}

The absolute difference and mutual information between weights of all convolutional layers of an original network and its fine-tuned offspring is reported in Figure \ref{fig:r_w_a}. Refer to the Section 4.2 of the principal manuscript for relevant discussion.

\begin{figure*}[ht]
    \centering
    \begin{tabular}{cc}
        \includegraphics[width=0.45\textwidth]{weights.png} & \includegraphics[width=0.45\textwidth]{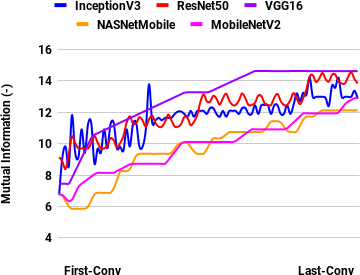}
    \end{tabular}
    \caption{Left panel: the \textbf{absolute difference} of kernel weights between an original network and its fine-tuned offspring. Right panel: the \textbf{mutual information} between kernels weights of an original network and its fine-tuned offspring}
    \label{fig:r_w_a}
\end{figure*}

\subsection{Activation maps comparison}

\begin{figure*}[ht]
    \centering
    \includegraphics[width=0.93\textwidth]{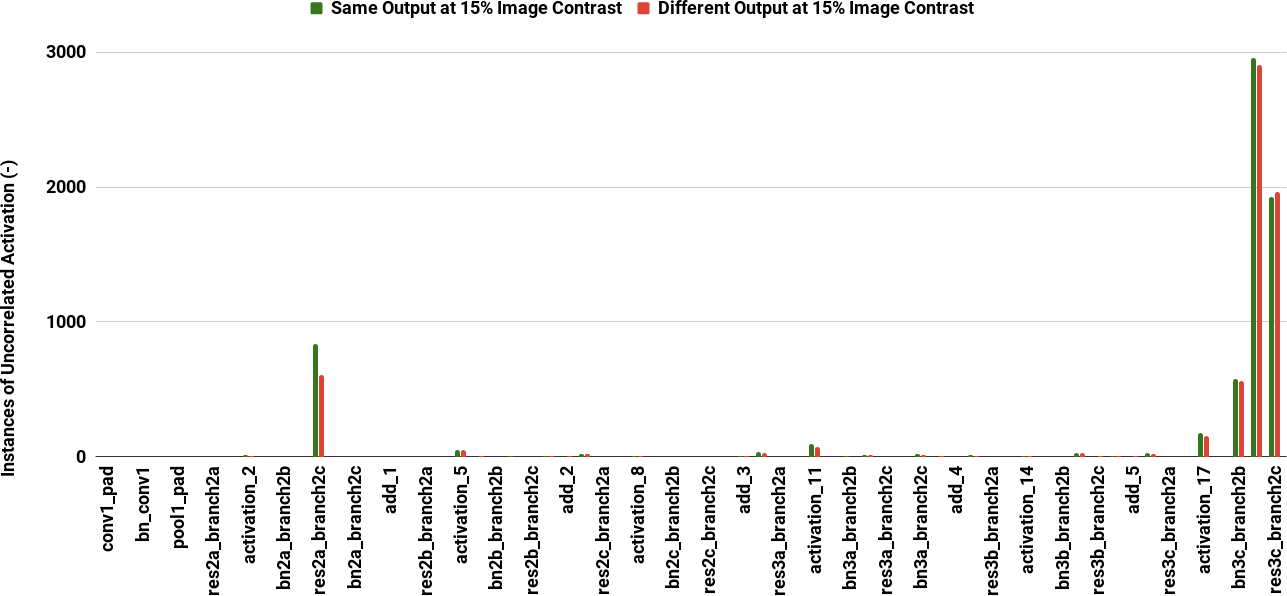}
    \caption{The total number of images in validation-set of \textit{ImageNet} with uncorrelated kernels' activation (\ie $corr \leq 0.75$) between \textit{ResNet50} and its contrast-augmented fine-tuned offspring. \textbf{Green} bars correspond to the images where both networks have a correct output. \textbf{Red} bars correspond to the instances where the fine-tuned network is successful while the original network fails.}
    \label{fig:actres}
\end{figure*}

Figure \ref{fig:actres} corresponds to the red bars of Figure 8 in the principal manuscript divided into two categories: (i) the green bars are the average of all images where the original \textit{ResNet50} and its fine-tuned offspring both correctly classify an image at 15\% level of contrast, (ii) red bars refer to those instances where the original \textit{ResNet50} fails to correctly classify while the fine-tuned version does classify correctly. As it can be observed there is no clear difference between the two set of bars. This suggests that although ``\textit{branch\_2c}" convolutional layers clearly exhibit a different behaviour in comparison to the other layers, they cannot explain the difference between successful and failure trials. This is a necessary condition and should be answered in future works.

\begin{figure*}[ht]
    \centering
    \includegraphics[width=0.93\textwidth]{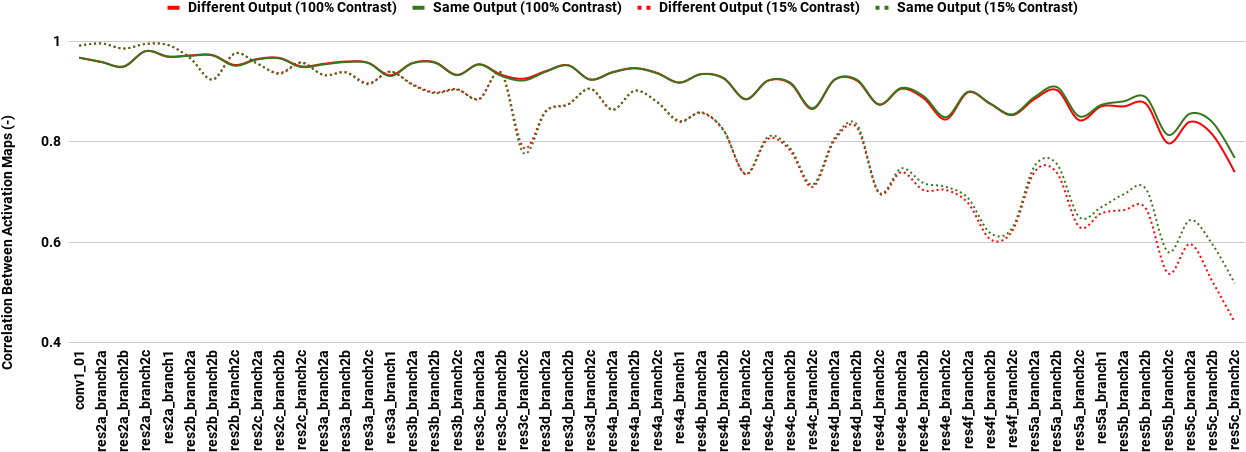}
    \caption{Correlation between activation maps of \textit{ResNet50} and its contrast-augmented fine-tuned offspring for all layer. The \textbf{continuous} lines correspond to images at 100\% level of contrast. The \textbf{dashed} lines correspond to images at 15\% level of contrast. \textbf{Green} colour refers to the images where both networks have a correct output. \textbf{Red} colour refers to the instances where the fine-tuned network is successful while the original network fails.}
    \label{fig:allact}
\end{figure*}

Figure \ref{fig:allact} shows the correlation between activation maps of the all layers in the original \textit{ResNet50} and its fine-tuned offspring. As it can be observed the green and red lines are very similar. This suggests correlation between activation maps of these twin networks cannot explain the difference between them.

\subsection{Successful versus failure comparison}

\begin{figure*}[ht]
    \centering
    \includegraphics[width=0.93\textwidth]{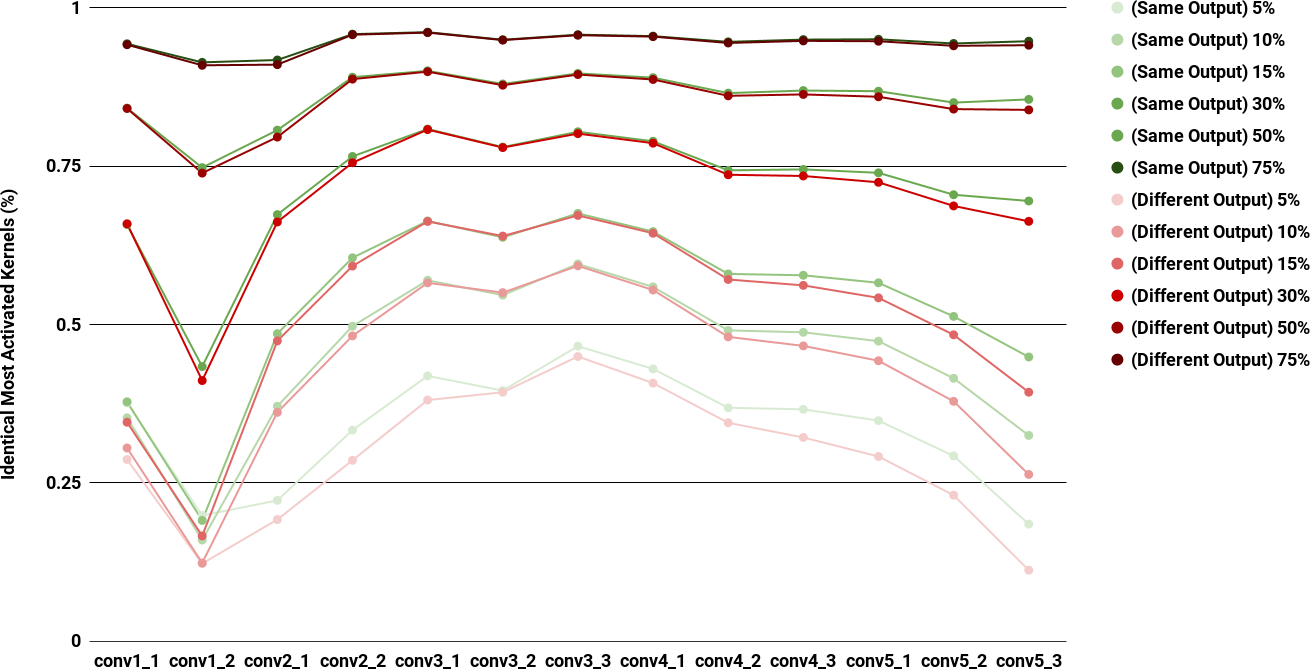}
    \caption{Comparison of the most activated kernels in activation maps of \textit{VGG16} between images at lower levels of contrast to 100\% contrast. \textbf{Green} colour correspond to the images where \textit{VGG16} is correct at both 100\% image contrast and lower image contrast. \textbf{Red} colour corresponds to the images where \textbf{VGG16} is correct at 100\% image contrast but it fails at a lower image contrast.}
    \label{fig:vgg16}
\end{figure*}

Figure \ref{fig:vgg16} shows the percentage of the most activated kernels that remain identical at every convolutional layer when the image contrast is reduced from 100\% to a lower level. As it can be observed from this figure, as contrast of image is reduced the percentage of identical kernels to 100\% image contrast is reduced. This can explain why \textit{VGG16} performs worse when contrast of an image is poor. However there is no difference between green and red lines regardless of the contrast examined. This suggests similar to above that percentage of the most activated kernels cannot explain why a network fails for certain images at low contrast, while it succeeds for others. This should be studied in future works.

\end{document}